\documentclass[10pt,twocolumn,letterpaper]{article}

\usepackage{iccv}
\usepackage{times}
\usepackage{epsfig}
\usepackage{graphicx}
\usepackage{amsmath}
\usepackage{amssymb}

\usepackage{lipsum}
\usepackage{booktabs}
\usepackage{subcaption}
\usepackage{paralist}
\usepackage{adjustbox}
\usepackage[sort,nocompress]{cite}
\usepackage[accsupp]{axessibility}  

\usepackage[pagebackref=true,breaklinks=true,colorlinks,bookmarks=false]{hyperref}

\DeclareGraphicsExtensions{.pdf,.jpg,.png}
\graphicspath{{figs/}}

\newcommand{\secref}[1]{Section~\ref{sec:#1}}
\newcommand{\figref}[1]{Figure~\ref{fig:#1}}
\newcommand{\tabref}[1]{Table~\ref{tbl:#1}}

\iccvfinalcopy 

\def\iccvPaperID{5871} 

\ificcvfinal\fi

\begin{document}

\title{Augmenting Depth Estimation with Geospatial Context}

\author{
  \centering
  \begin{minipage}{.9\linewidth}
    \centering
    \begin{minipage}{2.0in}
      \centering
      Scott Workman
    \end{minipage}
    \begin{minipage}{2.0in}
      \centering
      Hunter Blanton
    \end{minipage}
    \\[.2cm]
    \begin{minipage}{2.0in}
      \centering
      DZYNE Technologies
    \end{minipage}
    \begin{minipage}{2.0in}
      \centering
      University of Kentucky
    \end{minipage}
  \end{minipage}
}

\maketitle

\ificcvfinal\thispagestyle{empty}\fi

\begin{abstract}
    Modern cameras are equipped with a wide array of sensors that
    enable recording the geospatial context of an image. Taking
    advantage of this, we explore depth estimation under the
    assumption that the camera is geocalibrated, a problem we refer to
    as geo-enabled depth estimation. Our key insight is that if
    capture location is known, the corresponding overhead viewpoint
    offers a valuable resource for understanding the scale of the
    scene. We propose an end-to-end architecture for depth estimation
    that uses geospatial context to infer a synthetic ground-level
    depth map from a co-located overhead image, then fuses it inside
    of an encoder/decoder style segmentation network. To support
    evaluation of our methods, we extend a recently released dataset
    with overhead imagery and corresponding height maps. Results
    demonstrate that integrating geospatial context significantly
    reduces error compared to baselines, both at close ranges and when
    evaluating at much larger distances than existing benchmarks
    consider.
\end{abstract}

\section{Introduction}

Accurately estimating depth is important for applications that seek to
interpret the 3D environment, such as augmented reality and autonomous
driving. The traditional geometric approach for solving this problem
requires multiple views and infers depth by triangulating image
correspondences. Lately, more attention has been paid to the
single-image variant, which has great potential value but is known to
be ill-posed. Ranftl et al.~\cite{ranftl2020towards} point out that to
solve this problem ``one must exploit many, sometimes subtle, visual
cues, as well as long-range context and prior knowledge."

One of the primary difficulties with inferring depth from a single
image is that there is an inherent scale ambiguity. In other words,
different sized objects in the world can have the same projection on
the image plane (simply by adjusting the focal length or position in
space). Despite this, methods that take advantage of convolution
neural networks have shown promise due to their ability to capture
prior information about the appearance and shape of objects in the
world. 

\setlength{\fboxsep}{0pt}
\setlength{\fboxrule}{1px}

\begin{figure}
    \centering
    
    \fbox{\includegraphics[width=.234\linewidth]{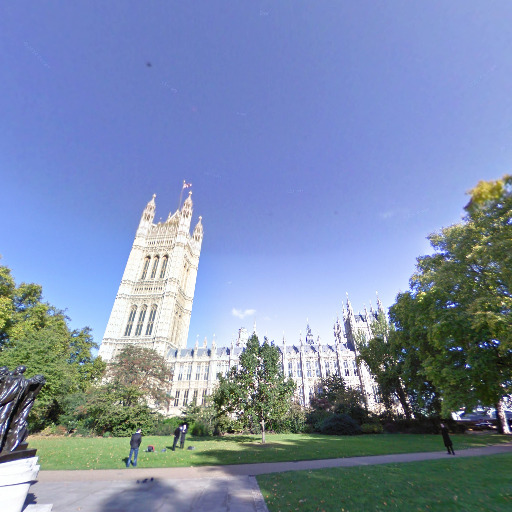}}
    \fbox{\includegraphics[width=.234\linewidth]{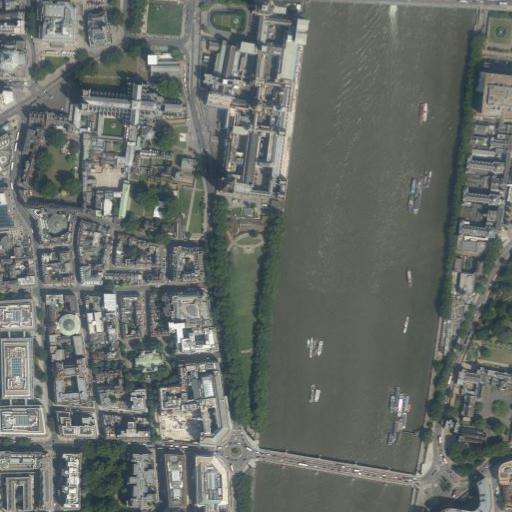}}
    \fbox{\includegraphics[width=.234\linewidth]{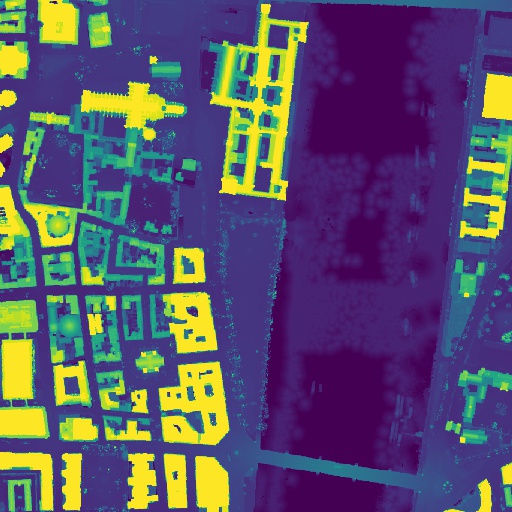}}
    \fbox{\includegraphics[width=.234\linewidth]{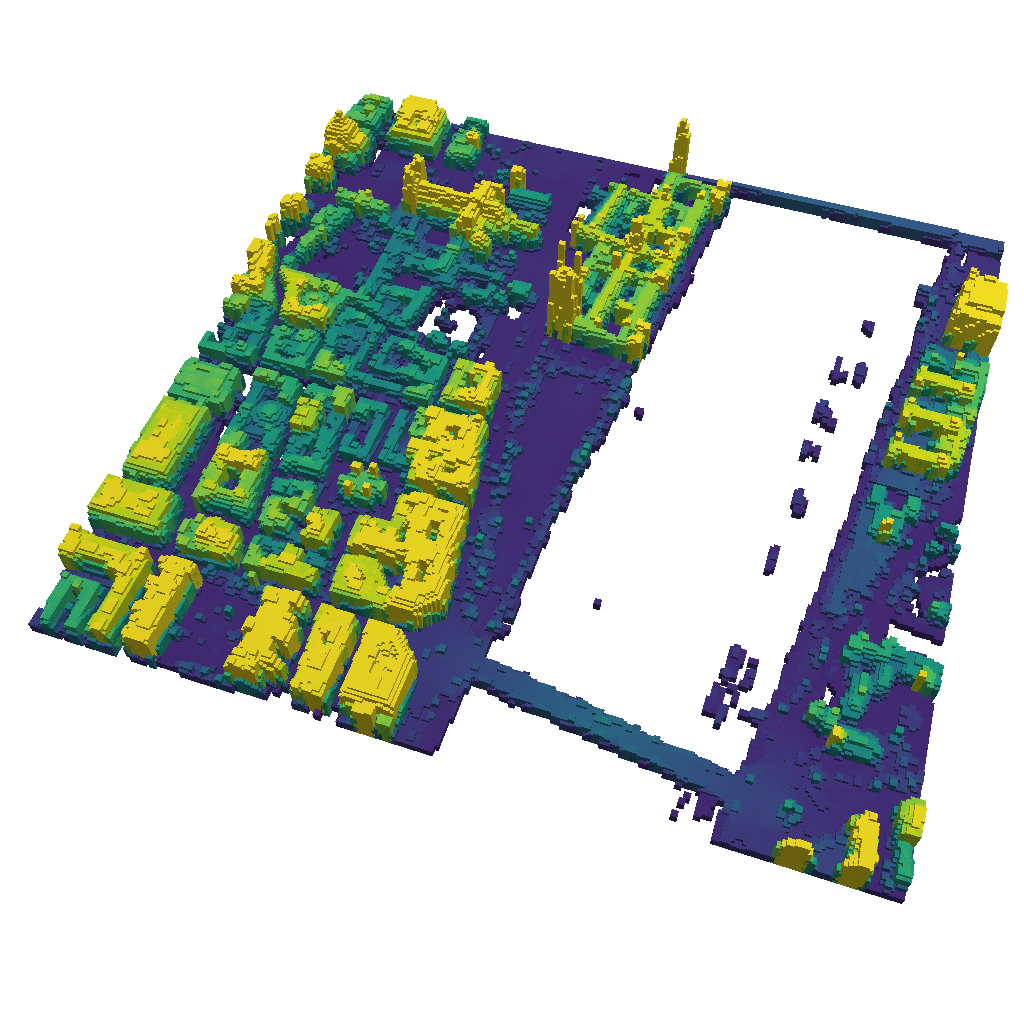}}
    \fbox{\includegraphics[width=.9915\linewidth,trim=0cm 1.5cm 0cm 1cm,clip]{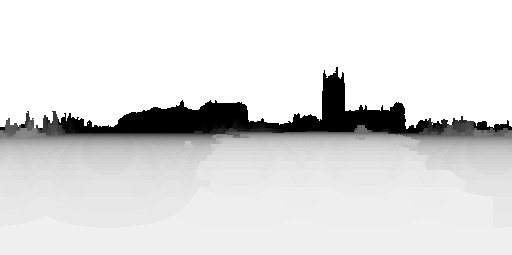}}
    
    \caption{We explore a new problem, {\em geo-enabled depth
    estimation}, in which the geospatial context of a query image is
    exploited during the depth estimation process.}
    \label{fig:cartoon}
\end{figure}

There are broadly two classes of methods in this space. Supervised
approaches assume a ground-truth labeling is provided during training,
often obtained from another sensor such as LiDAR. This labeling could
be absolute (metric values) or have an unknown scale. Self-supervised
approaches, on the other hand, do not require ground-truth depth.
Instead, the consistency of multiple inputs (e.g., sequences of images
from a video, or a stereo pair) are used to derive depth up to a
scaling factor, often by formulating the problem as a novel view
synthesis task. For both of these classes of methods, it is common to
make strong assumptions about the scale of the scene during training,
or to require computation of a scaling factor at inference time in
order to interpret the predicted depths.

For example, supervised methods often presume to know the maximum
observed depth of the scene, by constraining the output of the network
using a sigmoid activation and scaling by the maximum
depth~\cite{fu2018deep,lee2019big}. If the scale is unknown, i.e., a
scale-invariant loss was used during training, then a scaling factor
must be computed at inference to interpret the predictions relative to
the world. Such objective functions have been proposed when metric
depth is not available or for combining training datasets with
different properties. For example, Ranftl et
al.~\cite{ranftl2020towards} align their predictions with the
ground-truth via a least squares criterion before computing error
metrics. These caveats limit the generalizability of such methods when
applying them to real-world imagery from novel locations (e.g.,
varying depth ranges or lack of ground truth).

A similar phenomena occurs in self-supervised monocular approaches
that estimate depth up to an unknown scale. The maximum observed depth
of the scene is often used to constrain the predicted depths during
training, and a scaling factor is computed at inference to bring the
predictions in line with the ground truth. As before, the common
strategy in the current literature is to compute this scaling factor
using the ground-truth directly (per image), in this case by computing
the ratio of the median predicted values and median ground-truth
values~\cite{godard2019digging}. The issue of how to calibrate
self-supervised monocular depth estimation networks has only recently
been highlighted by McCraith et al.~\cite{mccraith2020calibrating},
who point out that current approaches severely limit practical
applications.

Beyond these issues, estimating depth at long ranges is known to be
extremely challenging. Zhang et al.~\cite{zhang2020depth} note the
limitations of LiDAR (sparse, reliable up to 200m) and argue the need
for ``dense, accurate depth perception beyond the LiDAR range." Most
state-of-the-art depth estimation networks assume a maximum depth of
100 meters for outdoor scenes~\cite{godard2019digging}. Further,
popular benchmark datasets for depth estimation are constrained to
small ranges, typically below 100 meters (using a {\em depth cap} to
filter pixels in the ground truth). For example, Ranftl et
al.~\cite{ranftl2020towards} evaluate on four different datasets,
ETH3D, KITTI, NYU, and TUM, with the depth caps set to 72, 80, 10, and
10 meters, respectively. Reza et al.~\cite{reza2018farsight} have
similarly pointed out the need for depth estimation to function at
much larger distances. 

In this work we explore how geospatial context can be used to augment
depth estimation, a problem we refer to as {\em geo-enabled depth
estimation} (\figref{cartoon}). Modern cameras are commonly equipped
with a suite of sensors for estimating location and orientation. Kok
et al.~\cite{kok2017using} provide an in-depth overview of algorithms
for recovering position/orientation from inertial sensors, concluding
that as quality has improved and cost has decreased ``inertial sensors
can be used for even more diverse applications in the future."
Accordingly, a great deal of work has shown that geo-orientation
information is extremely valuable for augmenting traditional vision
tasks~\cite{luo2008event,tang2015improving,mattyus2016hd,workman2017unified,workman2017natural}.

Given a geocalibrated camera, we explore how to inject geospatial
context into the depth estimation process. In this scenario, our goal
is to develop a method that takes advantage of the known
geocalibration of the camera to address the previously outlined
weaknesses. Specifically, we want to use geospatial context to 1)
reduce the inherent scale ambiguity and to 2) enable more accurate
depth estimation at large distances. Our key insight is that if the
location of the capturing device is known, the corresponding overhead
viewpoint is a valuable resource for characterizing scale.

We propose an end-to-end architecture for depth estimation that uses
geospatial context to infer an intermediate representation of the
scale of the scene. To do this, we estimate a height (elevation) map
centered at the query image and transform it to a synthetic
ground-level depth map in a differentiable manner via a sequence of
voxelization and ray casting operations. This intermediate
representation is metric, and we fuse it inside of an encoder/decoder
segmentation architecture that outputs absolute depth estimates.
Importantly, our approach makes no assumptions during training about
the maximum observed depth and requires no post-processing step to
align predictions.

To support evaluating our methods, we extend the recently released
HoliCity dataset~\cite{zhou2020holicity} to include overhead imagery
and corresponding height data from a composite digital surface model.
Extensive experiments show that when geospatial context is available
our approach significantly reduces error compared to baselines,
including when evaluating at much longer depth ranges than considered
by previous work. 

\begin{figure*}

  \centering

  \begin{subfigure}{1\linewidth}
    \centering
    
    \includegraphics[width=.325\linewidth]{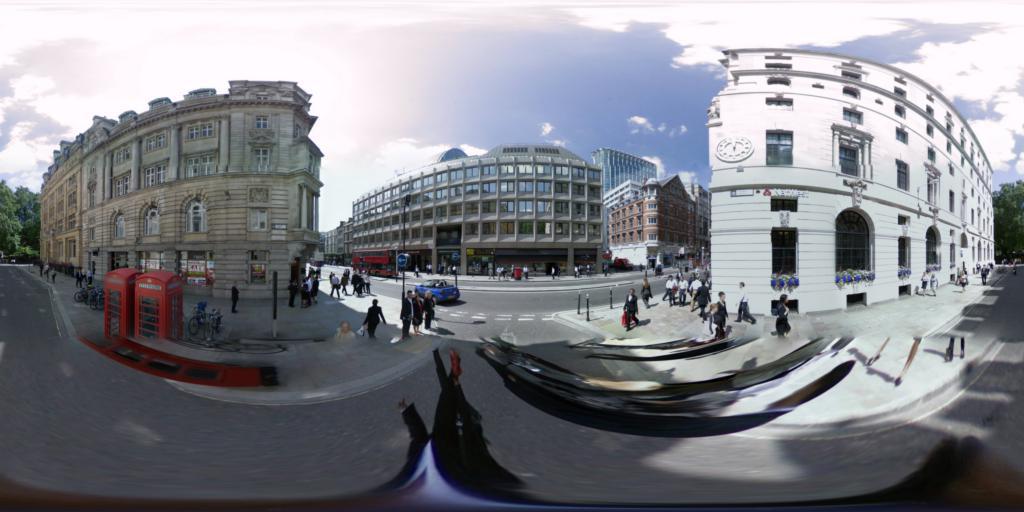}
    \includegraphics[width=.1625\linewidth]{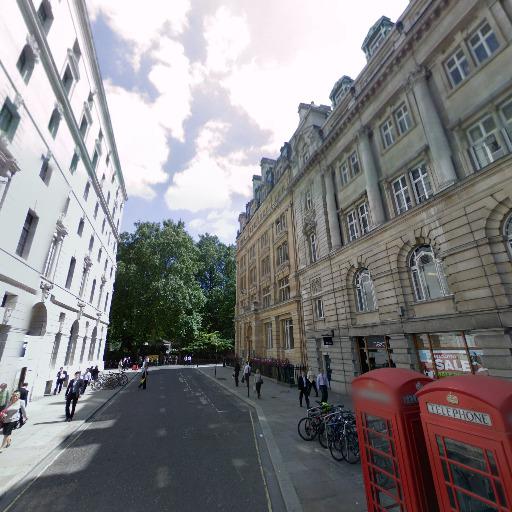}
    \includegraphics[width=.1625\linewidth]{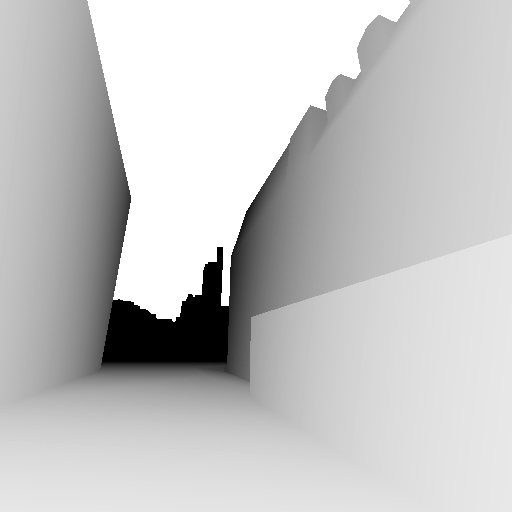}
    \includegraphics[width=.1625\linewidth]{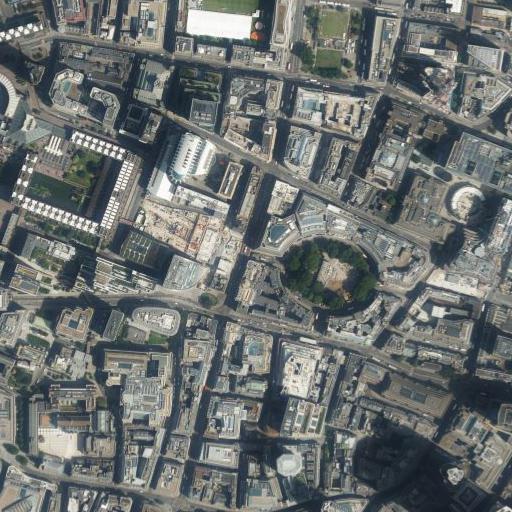}
    \includegraphics[width=.1625\linewidth]{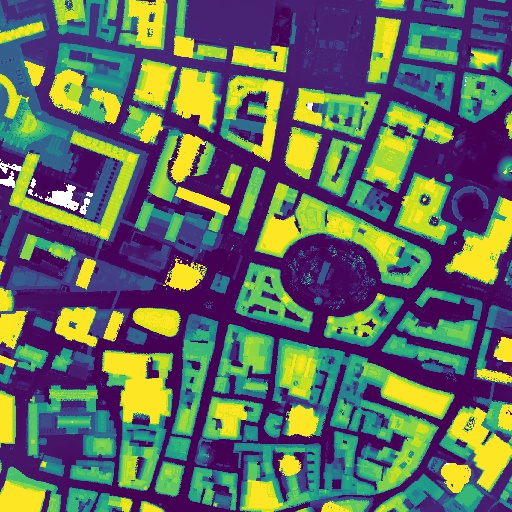}

    \includegraphics[width=.325\linewidth]{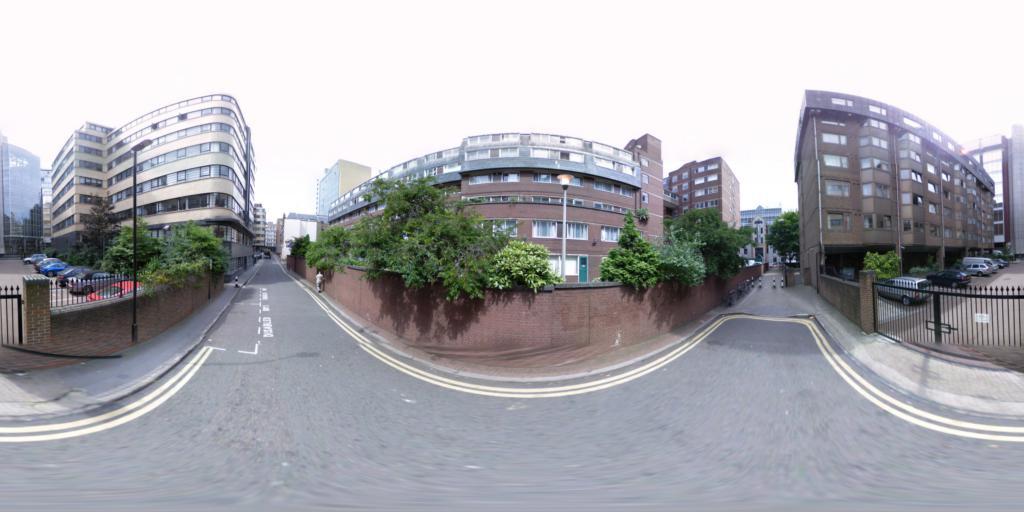}
    \includegraphics[width=.1625\linewidth]{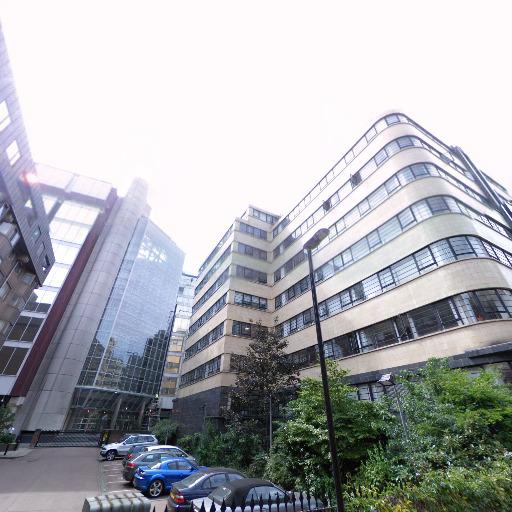}
    \includegraphics[width=.1625\linewidth]{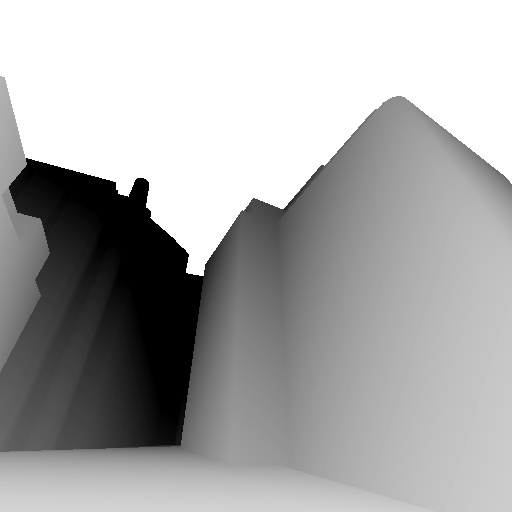}
    \includegraphics[width=.1625\linewidth]{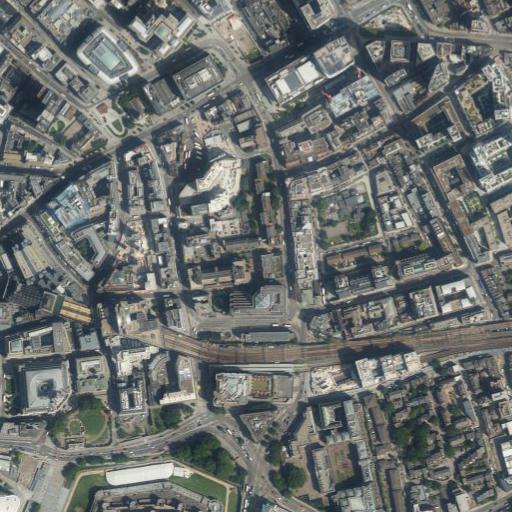}
    \includegraphics[width=.1625\linewidth]{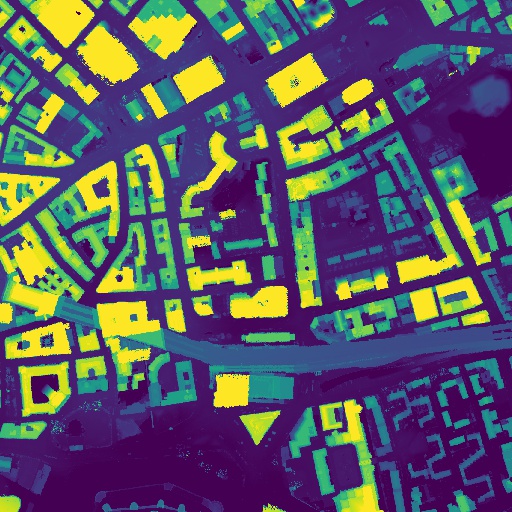}
  \end{subfigure}

  \caption{We introduce the {\em HoliCity-Overhead} dataset which
  extends the recently introduced HoliCity
  dataset~\cite{zhou2020holicity} to include overhead imagery and
  associated ground-truth height maps. From left to right,
  equirectangular panorama, perspective cutout, corresponding depth
  map, co-located overhead image, and corresponding height map.}
  
  \label{fig:dataset}
\end{figure*}

\section{Related Work}

Traditional work in depth estimation relied on geometric cues from
multiple images to infer depth. Interest quickly shifted to the
single-image variant of the problem with early approaches relying on a
set of assumptions about the geometric layout of the
scene~\cite{hoiem2005automatic}. For example, Delage et
al.~\cite{delage2006dynamic} proposed a 3D reconstruction method for a
single indoor image that makes assumptions about the relationship
between vertical and horizontal surfaces and uses visual cues to find
the most probable floor-wall boundary. Saxena et
al.~\cite{saxena2008make3d} later assumed the environment is made up
of many small planes and estimated the position and orientation of
each using a Markov random field.

More recently in machine vision it has become common to directly
regress depth using convolutional neural networks. Supervised
approaches use ground-truth depth from RGB-D cameras, LiDAR sensors,
or stereo matching~\cite{eigen2014depth}. In this space there has been
much exploration into various architecture and design
choices~\cite{laina2016deeper,fu2018deep,alhashim2018high,lee2019big,
lee2019monocular}. However, the primary challenge for supervised
methods remains the difficulty in acquiring high quality and varied
training data. To navigate this issue, Atapour-Abarghouei and
Breckon~\cite{atapour2018real} propose to train using a synthetic
dataset and then apply style transfer to improve performance on
real-world images. Other work has relaxed the requirement for absolute
depth supervision by proposing scale-invariant objective
functions~\cite{chen2016single}. Ranftl et
al.~\cite{ranftl2020towards} argue that performance is primarily
impacted by the lack of large-scale ground truth, proposing a
scale-invariant loss that enables mixing of data sources.
 
Alternatively, self-supervised methods circumvent the need for
ground-truth depth entirely, instead relying on multiple inputs (e.g.,
sequences of images from a video, or a stereo pair) during training to
derive depth up to a scaling factor. The problem is commonly
reformulated as an image reconstruction task
with~\cite{godard2017unsupervised} or
without~\cite{zhou2017unsupervised,godard2019digging, zhao2020towards}
known camera pose information. While self-supervised methods that take
advantage of stereo supervision can infer scale directly from the
known camera baseline~\cite{godard2019digging}, self-supervised
monocular approaches suffer from the need to align predictions with
the ground-truth at inference by computing the scaling
factor~\cite{mccraith2020calibrating}. When considering both
supervised and self-supervised approaches, it is common to make an
assumption about the maximum observed depth during
training~\cite{fu2018deep,lee2019big}. In addition, popular benchmarks
such as ETH3D and KITTI are evaluated sub-100 meters. This limits the
practical application of these methods when considering images from
novel locations, thus methods that function at larger distances are
needed~\cite{reza2018farsight}.

Motivated primarily by the surge in interest in autonomous driving,
another strategy is to frame the problem as depth completion or depth
refinement where, in addition to the input image, an approximate
(possibly sparse) depth image is provided (e.g., from a LiDAR sensor).
Here, the objective is to produce a dense, more accurate depth
map~\cite{rossi2020joint}. Our approach is similar to this line of
work in the sense that we use geospatial context to produce an
intermediate depth estimate that is used along with the input image to
infer a final depth prediction. Though we focus on integrating
geospatial context, our method can conceivably be combined with any
recent depth refinement approach.

Geospatial context has become a powerful tool for improving the
performance of traditional vision tasks. For example, Tang et
al.~\cite{tang2015improving} consider the task of image classification
and show how geolocation can be used to incorporate several different
geographic features, ultimately improving classification performance.
Similarly, overhead imagery has proven to be useful as a complementary
viewpoint of the scene. Luo et al.~\cite{luo2008event} combine
hand-crafted features for a pair of ground-level and overhead images
to improve ground-level activity recognition. In the realm of image
geolocalization, overhead imagery has been used as an alternative to a
ground-level reference database~\cite{lin2013cross,workman2015wide} to
enable dense coverage. Other use cases include making maps of
objects~\cite{mattyus2016hd,wegner2016cataloging} and visual
attributes~\cite{workman2017natural,salem2020learning}, understanding
traffic patterns~\cite{workman2020dynamic}, detecting
change~\cite{ghouaiel2016coupling}, and visualizing
soundscapes~\cite{salem2018soundscape}. To our knowledge, this work is
the first to consider how geospatial context can be used to improve
depth estimation.

\begin{figure*}
    \centering
    \includegraphics[width=1\linewidth]{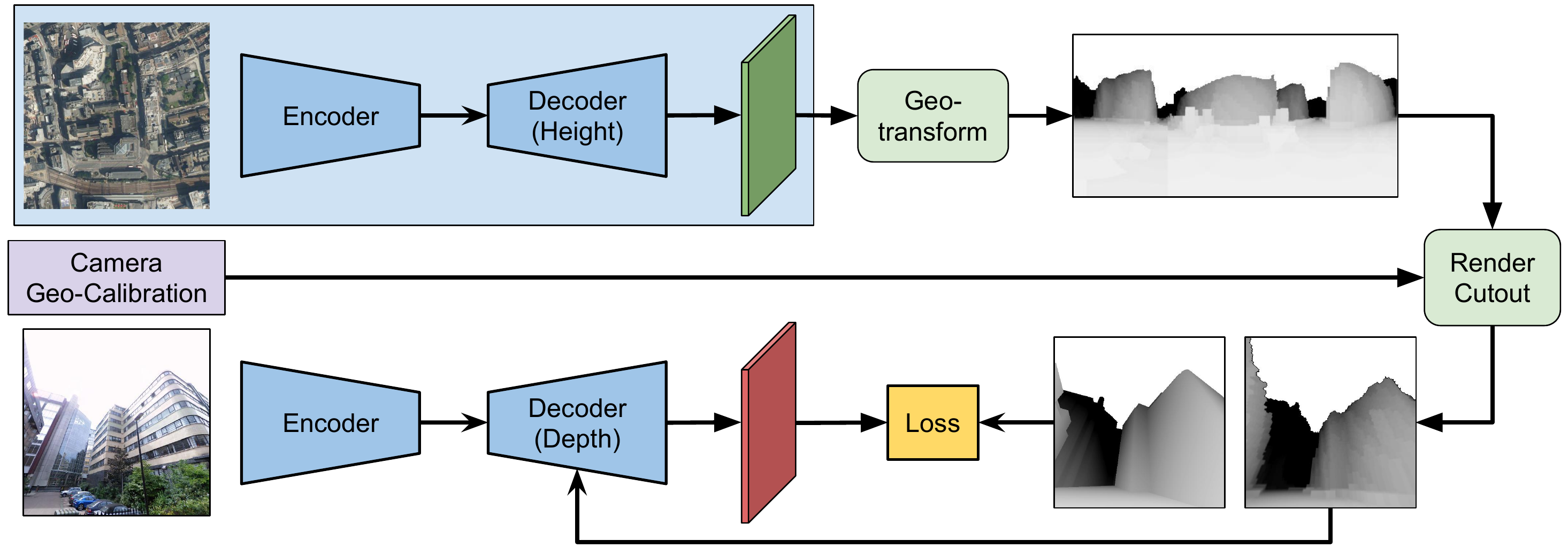}
    \caption{An overview of our approach. Given a geolocated image, we
    transform a co-located height map to an intermediate
    representation of the scale of the scene via a series of
    differentiable operations that take advantage of the known camera
    geocalibration. We then fuse it into an encoder/decoder
    segmentation architecture that operates on the ground-level image.
    Importantly, our approach can function on a known height map if
    available (e.g., from a composite DSM), or instead estimate height
    from a co-located overhead image (shaded region).}
    \label{fig:architecture}
\end{figure*}

\section{HoliCity-Overhead Dataset}

To support our experiments, we introduce the HoliCity-Overhead dataset
which extends the recently introduced HoliCity~\cite{zhou2020holicity}
dataset. HoliCity is a city-scale dataset for learning holistic 3D
structures such as planes, surface normals, depth maps, and vanishing
points. The dataset was constructed by taking advantage of a
proprietary computer-aided design (CAD) model of downtown London,
United Kingdom with an area of more than $20km^2$. Note that as labels
are derived from the CAD model, they do not contain dynamic objects
(e.g., pedestrians). We do not consider this a limitation as deriving
depth in this manner enables ground-truth depth values at
significantly greater ranges compared to existing datasets (on the
order of kilometers for HoliCity), which is crucial for supporting our
goal of enabling more accurate depth estimation at larger distances.

In the source region, 6,300 panoramas were collected from Google
Street View with a native resolution of $6,656 \times 13,312$. The
individual panoramas were aligned with the CAD model such that the
average median reprojection error is less than half a degree. From
each panorama, eight perspective cutouts (of size $512 \times 512$)
were extracted at 45 degrees apart, with yaw and pitch angles randomly
sampled and field-of-view set to $90^\circ$. Labels were generated for
each cutout using the CAD model. Importantly, the geo-orientation
information for the original $360^{\circ}$ equirectangular panoramas,
as well as the camera parameters defining the perspective cutouts, are
provided. Example images from the dataset are shown in
\figref{dataset}. 

For our purposes, we extended the dataset to include overhead imagery
and ground-truth height maps, which we refer to as the {\em
HoliCity-Overhead} dataset. For each Google Street View panorama, we
collected a co-located overhead image at multiple resolutions (zoom
levels 16-18) from Bing Maps (each of size $512 \times 512$). Then, we
generated a height map for each overhead image by aligning to a 1
meter composite digital surface model (DSM) of London produced by the
Environment Agency in 2017. The DSM data is made publicly available
via the UK government at the open data
portal.\footnote{\url{https://data.gov.uk/}} Examples of the resulting
overhead image and height map pairs contained in the HoliCity-Overhead
dataset are shown in \figref{dataset} (right). 

Though HoliCity provides an official evaluation split, ground-truth
data for the test set is reserved for a future held-out benchmark. As
such, in our experiments we report performance numbers using the
validation set and instead reserve a small portion of the training set
for validation. 

\section{Geo-Enabled Depth Estimation}
\label{sec:method}

We propose an end-to-end architecture for depth estimation that
integrates geospatial context. \figref{architecture} provides a visual
overview of our approach. For the purposes of description, we outline
our approach as if a height map is estimated from a co-located
overhead image, but it can be provided directly as input if available.

\subsection{Approach Overview}

Given a geocalibrated ground-level image (i.e., known geolocation,
orientation, field of view), our approach has two primary components.
First, we estimate a height map from a co-located overhead image and
use it to generate an intermediate representation of the scale of the
scene. To generate the intermediate representation from the height
map, we render a synthetic ground-level depth image through a sequence
of differentiable operations that take advantage of the known camera
geocalibration (i.e., conversion to a voxel representation and ray
casting). This intermediate representation is metric and has many
potential uses. The second component of our approach performs joint
inference on a ground-level image and the synthetic depth image in an
encoder/decoder style segmentation architecture, fusing the two
modalities inside the decoder.

\subsection{Inferring Scale from an Overhead Viewpoint}

We leverage geospatial context to generate an intermediate
representation of the scale of the scene from an overhead viewpoint. 

\subsubsection{Estimating a Height Map}

Given a geocalibrated ground-level image and a co-located overhead
image, we first estimate a per-pixel height map from the overhead
image. We represent this as a supervised regression task that outputs
per-pixel metric height values. For our segmentation architecture, we
use LinkNet~\cite{chaurasia2017linknet} with a
ResNet-34~\cite{he2016deep} encoder (initialized using weights from a
model trained on ImageNet~\cite{deng2009imagenet}). For the objective
function, we minimize the Pseudo-Huber loss (also recognized as the
Charbonnier loss):
\begin{equation}
    \mathcal{L}_{height} = \delta^2 (\sqrt{1 + ((y-\hat{y})/\delta)^2} - 1),
    \label{eq:pseudo-huber}
\end{equation}
where $y$ and $\hat{y}$ are the observed and predicted values,
respectively. The Pseudo-Huber loss is a smooth approximation of the
Huber loss, where $\delta$ controls the steepness.

\subsubsection{Synthesizing a Depth Panorama}

\begin{figure}

  \centering

  \begin{subfigure}{.325\linewidth}
    \centering
    \includegraphics[width=1\linewidth]{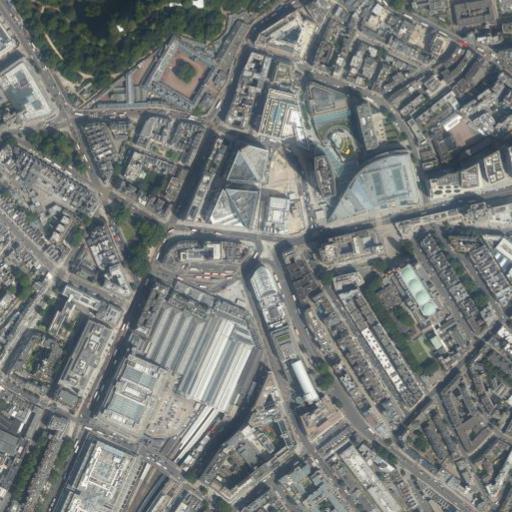}
    \includegraphics[width=1\linewidth]{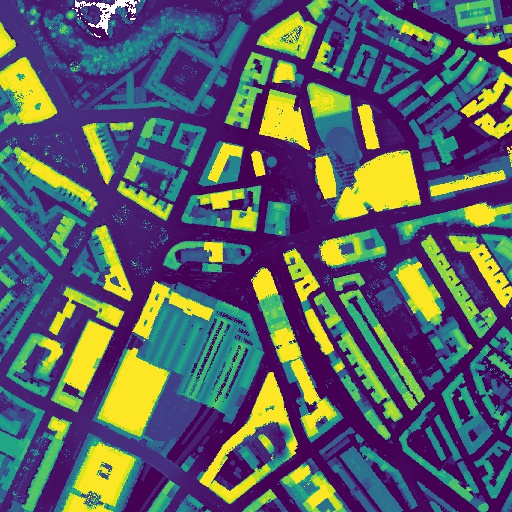}
  \end{subfigure}
  \begin{subfigure}{.654\linewidth}
    \includegraphics[width=1\linewidth]{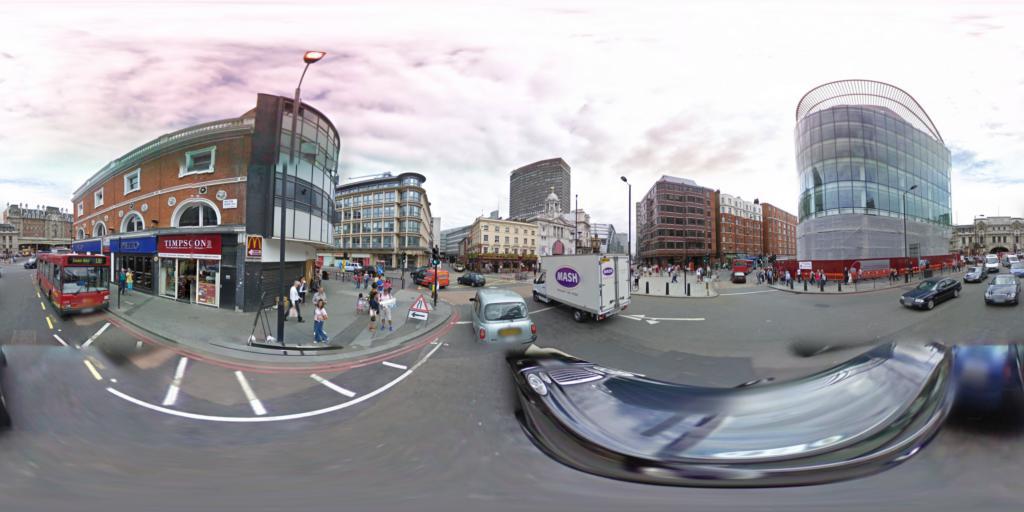}
    \includegraphics[width=1\linewidth]{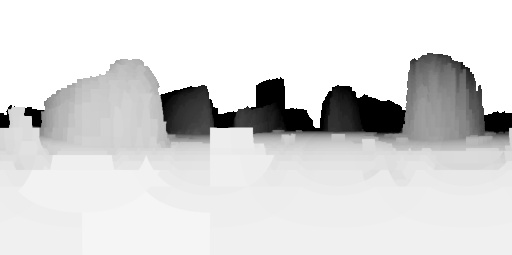}
  \end{subfigure}

  \caption{Transforming a height map from an overhead viewpoint to a
  depth panorama using a voxel representation combined with ray
  casting. (left, bottom) The input height map and (right, bottom) the
  generated depth panorama. As the ground sample distance of the
  overhead height map is known, the resulting depth panorama is
  metric.}
  
  \label{fig:geotransform}
\end{figure}

Drawing inspiration from Lu et al.~\cite{lu2020geometry} who tackle
the problem of cross-view image synthesis, we use the estimated height
map to render north aligned panoramic depth images. Given that the
overhead imagery has known ground sample distance (the spatial extent
of each pixel in the world is known), we use the overhead height map
to construct a voxel occupancy grid. The grid is generated such that
voxel $v_{i,j,k} = 1$ if height value $h_{i,j} > k$ at pixel location
$(i,j)$. The overhead image, and subsequently the voxel grid, is
centered at the geolocation of the query ground-level image. Then, a
synthetic panoramic depth image is constructed from the voxel grid by
sampling at uniform distances along the ray for each pixel in the
output panorama. The output depth is set to the minimum sampling
distance that intersects a non-zero voxel. \figref{geotransform}
visualizes the output of this process using a ground-truth height map.

\subsubsection{Extracting a Perspective Cutout}

The previous step generates a synthetic ground-level panoramic depth
image directly from an overhead height map. For use in our end-to-end
system, we also implement a differentiable layer for extracting
perspective cutouts from a 360$^\circ$ panorama. Given an
equirectangular panorama and target geocalibration (yaw, pitch, roll,
field of view), we extract the corresponding perspective image by
treating the panorama as a cylindrical image and sampling the
projections onto the image plane under the given camera geometry. We
implement this as a separate layer so that the panoramic depth image
can be accessed directly, and additionally for resource conservation
in the event that several perspective cutouts are needed from a single
panorama. 

\subsection{Depth Refinement using Geospatial Context}

\begin{table*}
  \centering
  \caption{HoliCity evaluation results (depth cap 80m).}
  \begin{tabular}{@{}lccccccc@{}}
    \toprule
    & Abs Rel & Sq Rel & RMSE & RMSE log & $\delta < 1.25$ & $\delta < 1.25^2$ & $\delta < 1.25^3$ \\
    \midrule
    {\em overhead} & 0.886 & 21.564 & 10.571 & 0.602 & 0.314 & 0.600 & 0.771  \\
    {\em ground (variant of \cite{alhashim2018high})} & 0.503 & 10.631 & 5.626 & 0.301 & 0.675 & 0.876 & 0.940 \\
    {\em ours (concatenate)} & 0.515 & 14.790 & 5.057 & 0.290 & 0.711 & 0.883 & 0.944  \\
    {\em ours} & 0.501 & 12.205 & 4.895 & 0.279 & 0.711 & 0.893 & 0.950 \\
    \midrule
    {\em ground + median scaling (overhead)} & 0.923 & 25.355 & 8.635 & 0.457 & 0.400 & 0.709 & 0.847  \\
    {\em ground + median scaling (ground truth)} & 0.229 & 3.249 & 5.388 & 0.255 & 0.749 & 0.905 & 0.957 \\
    {\em ours + median scaling (ground truth)} & 0.216 & 2.967 & 4.782 & 0.239 & 0.774 & 0.920 & 0.964  \\
    
    \bottomrule
  \end{tabular}
  \label{tbl:holicity_results}
\end{table*}

Here we outline our depth refinement architecture
(\figref{architecture}, bottom) that takes as input a ground-level
image and the intermediate estimate of scale generated from a
co-located height map. We start from the architecture proposed by
Alhashim and Wonka~\cite{alhashim2018high} and regress depth using an
encoder/decoder segmentation network with skip connections. In this
approach, the decoder consists of a series of upsampling blocks. In
each block, the input feature map is upsampled via bilinear
interpolation, concatenated with the corresponding feature map from
the encoder (skip connection), and passed through two $3 \times 3$
convolutional layers with the number of output filters set to half of
the input filters. Unlike existing work, which often estimates half
resolution depth, we add an extra convolutional transpose layer before
the final output layer of the decoder in order to generate full
resolution depth. For the encoder, we use
DenseNet-161~\cite{huang2017densely} pretrained on ImageNet. 

To incorporate geospatial context (in the form of a synthetic depth
image obtained from the estimated height map) we fuse it with image
features inside the decoder. Specifically, before each convolutional
layer and upsampling block, we concatenate the synthetic depth image
as an additional channel of the input feature map, resizing as
necessary. The final two layers of the decoder (convolutional
transpose layer and output convolutional layer) are excluded from this
process. Fusing in the decoder allows the encoder to learn features
solely focusing on the content of the query image. 

Similar to height estimation, we minimize the Pseudo-Huber Loss
(\ref{eq:pseudo-huber}). However, we omit pixels from the objective
function that do not have ground-truth depths using a validation mask.
The final loss becomes
\begin{equation}
    \mathcal{L} = \alpha_{h}\mathcal{L}_{height} + \mathcal{L}_{depth},
\end{equation}
where $\alpha_{h}$ is a weighting term used to balance the two tasks.
Our approach can be thought of as a depth refinement technique that
takes into account the overhead approximation of scene scale.

\subsection{Implementation Details}

We implement our methods using PyTorch~\cite{paszke2019pytorch} and
PyTorch Lightning~\cite{falcon2019pytorch}. Our networks are optimized
using Adam~\cite{kingma2014adam} with the initial learning rate set to
$1e^{-4}$. All networks are trained for 25 epochs and the learning
rate policy is set to reduce-on-plateau by an order of magnitude using
the validation set (patience equal to 5 epochs). For the Pseudo-Huber
loss, we set $\delta=2$. To balance the two tasks, we set
$\alpha_{h}=0.1$. This weighting term is decreased a single time after
5 epochs, by a factor of 10. When estimating heights, we normalize
each ground-truth height map individually such that the minimum value
is zero. For rendering perspective cutouts, we set non-intersections
to a value of -1. 

\section{Evaluation}

We evaluate our methods quantitatively and qualitatively through a
variety of experiments. Results demonstrate that our approach, which
builds on a recent state-of-the-art method to inject geospatial
context, significantly reduces error at close ranges while
simultaneously enabling more accurate depth estimates at larger ranges
than have been previously considered.

\paragraph{Baseline Methods}

To evaluate the proposed architecture, we compare against several
baseline methods that share low-level components with our proposed
method. Our full approach is outlined in \secref{method} and is
subsequently referred to as {\em ours}. We also compare against a
baseline that omits geospatial context from our approach (referred to
as {\em ground}). Note that without geospatial context, this baseline
is simply a variant of the recent state-of-the-art method of Alhashim
and Wonka~\cite{alhashim2018high}. Additionally, we compare against a
baseline that uses only the intermediate estimate of scale, derived
from geospatial context, as the final prediction (referred to as {\em
overhead}). Finally, we compare against a baseline that concatenates
the intermediate estimate as an additional channel to the input image
and we refer to this as {\em ours (concatenate).} The strategy for
this baseline is similar in concept to the recent work of Liu and
Li~\cite{liu2019lending} who add orientation as an additional input
channel for cross-view image geolocalization. 

\subsection{Ablation Study}

We present results using the HoliCity-Overhead dataset. As mentioned
previously, we report metrics on the HoliCity~\cite{zhou2020holicity}
validation set as ground-truth data for the test set is unavailable.
Unless otherwise specified, all methods are trained using
HoliCity-Overhead data corresponding to zoom level 17  (approx. 0.74
meters per pixel, or 190 meter half width) and use the known height
map. 

\begin{table}
  \centering
  \caption{Estimated versus known height maps (HoliCity, depth cap 80m).}
  \begin{tabular}{@{}lcccc@{}}
    \toprule
    & RMSE & RMSE log \\
    
    \bottomrule
    {\em ours (estimated height)} & 4.935 & 0.287 \\
    {\em ours (known height)} & 4.895 & 0.279 \\
    
    \bottomrule
  \end{tabular}
  \label{tbl:learned_vs_known}
\end{table}

For our initial experiment, we evaluate the ability of our approach at
short ranges (depth cap of 80 meters), computing metrics as
in~\cite{godard2019digging}. \tabref{holicity_results} summarizes the
results of this study. As expected, the ground-only baseline
outperforms the overhead-only baseline, likely due to the difficulty
in precisely recovering fine-grained details from an overhead
viewpoint. Despite the limited evaluation range, our methods that
integrate geospatial context significantly outperform all baselines,
e.g., by over half a meter in RMSE versus the ground-only baseline.
Additionally, our approach of fusing in the decoder outperforms the
variant of our method that concatenates as an additional input
channel. 

In addition, we show the impact of adding median scaling (a per-image
scaling factor used to align results) to the ground-only baseline
using both the {\em overhead} estimate and the ground truth. This
result demonstrates the benefit of our end-to-end architecture over an
approach that simply uses the intermediate estimate of scale directly
as a calibration tool. Though we have previously noted the
impracticality of median scaling using the ground truth, for fairness
we show that our approach can similarly benefit, achieving
significantly lower error.

Finally, \tabref{learned_vs_known} shows that our method that
simultaneously learns height maps (from co-located overhead images)
performs competitively against our approach that accepts known height
maps directly (e.g., from a composite DSM). These results show that
geospatial context, if available, can be extremely useful for
augmenting depth estimation, even at small ranges.

\subsection{Long Range Depth Estimation}
Next, we analyze the performance of our methods at much greater
distances. One of the major limitations of existing work is that
evaluation is typically limited to less than 100
meters~\cite{ranftl2020towards,reza2018farsight}. This can be partly
attributed to the increased difficulty of accurately estimating depth
at long ranges, but also due to the limited range of LiDAR sensors,
which are often used for collecting ground truth. An advantage of the
HoliCity dataset~\cite{zhou2020holicity} is that the truth labels are
derived from a CAD model, enabling ground-truth depth to reflect much
larger distances.

\begin{figure}
    \centering
    \includegraphics[width=1\linewidth]{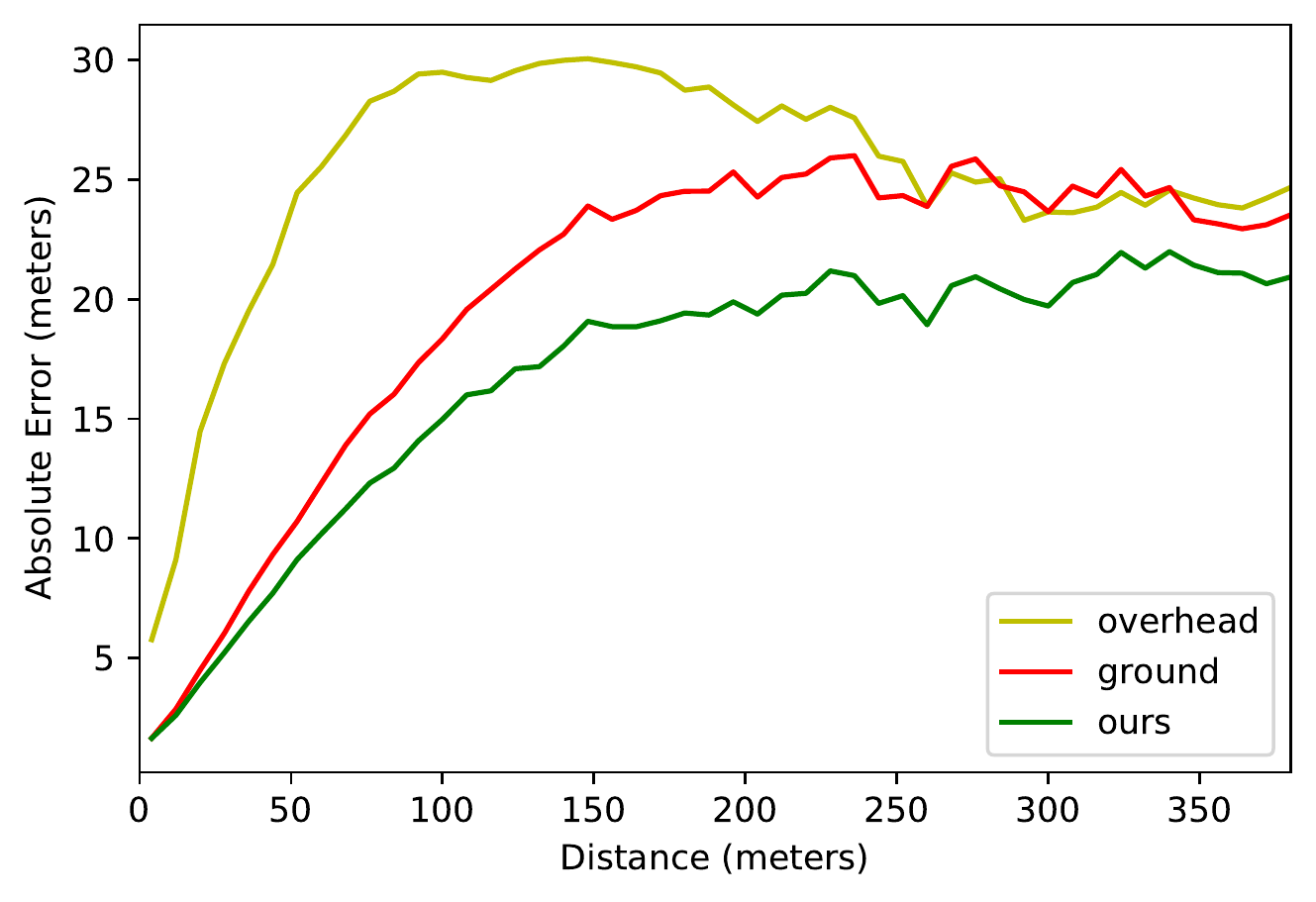}
    \caption{Integrating geospatial context reduces average error as
    distance increases when compared to baselines, including an
    overhead-only approach.}
    \label{fig:error_vs_distance}
\end{figure}

\figref{error_vs_distance} visualizes the performance of our approach
over a range of up to 400 meters, using absolute error as the metric,
versus two baselines. As expected, average error increases as the
magnitude of the depth increases. Our method not only exhibits lower
depth error overall, but greatly reduces error at long ranges. We
attribute this to our explicit intermediate representation of scale
derived from an overhead viewpoint, which enables a good approximation
of depth even at far distances.

\begin{figure}
    \centering
    \includegraphics[width=1\linewidth]{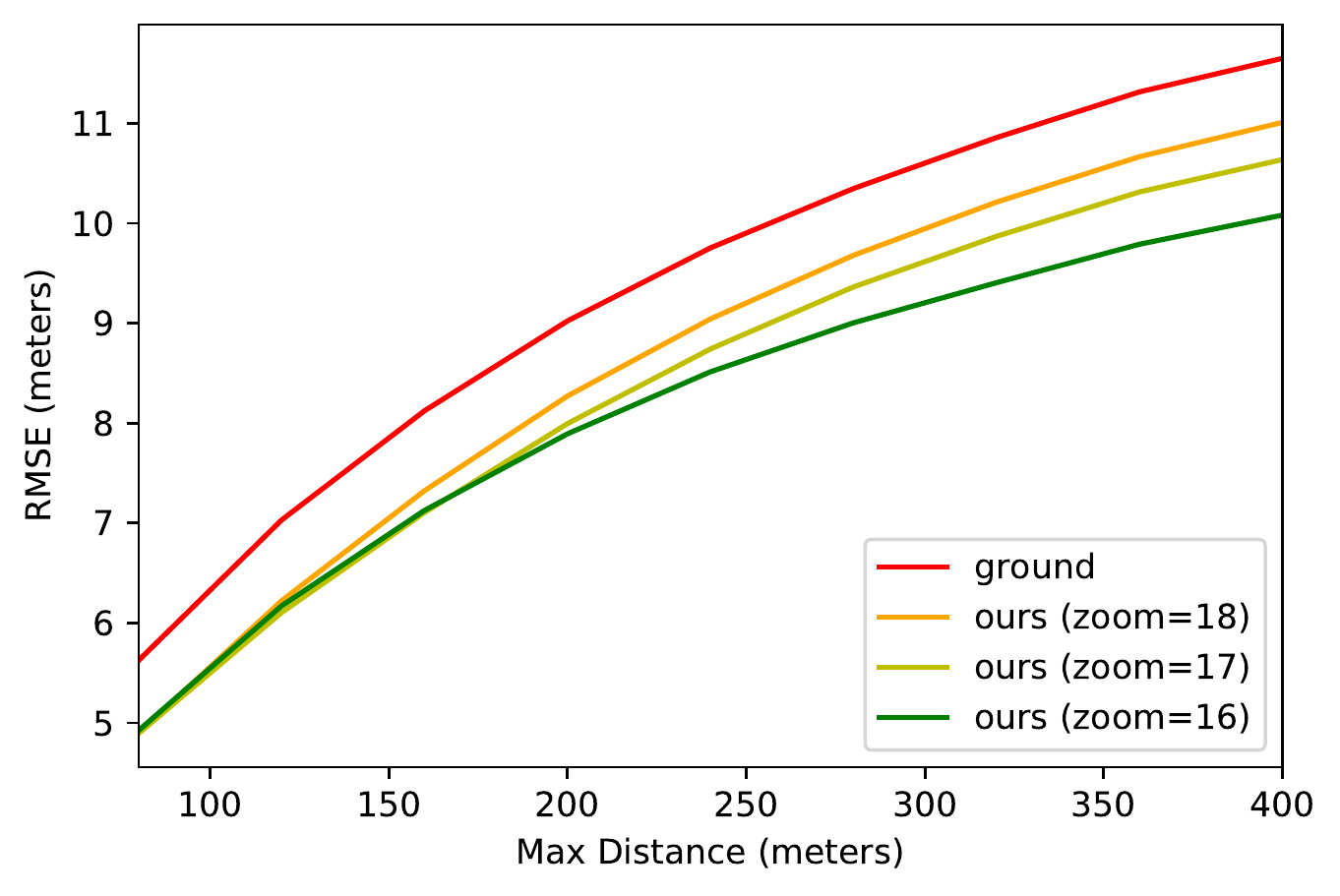}
    \caption{Evaluating the impact of varying ground sample distance.
    As anticipated, the greater spatial coverage of lower zoom levels
    positively impacts depth estimation performance at greater
    distances.}
    \label{fig:rmse_vs_distance}
\end{figure}

Finally, we evaluate the impact of varying ground sample distance on
our methods. In other words, does having a greater spatial coverage in
the overhead height map positively impact depth estimation
performance? Intuitively this makes sense, as greater spatial coverage
in the height maps would enable capturing objects further away in the
synthetic depth panorama (\figref{geotransform}) and subsequent
perspective cutout, with the trade-off of less detail (i.e., being
zoomed out). For this experiment, we train variants of our method for
the different zoom levels of imagery contained in HoliCity-Overhead.
\figref{rmse_vs_distance} visualizes the results, with the x-axis
representing the maximum depth considered (depth cap) when computing
the error metric (RMSE). As anticipated, at further distances,
starting from height maps with greater spatial coverage leads to an
advantage, with all methods significantly outperforming the
ground-only baseline.

\subsection{Impact of Geo-Orientation Accuracy}

\begin{figure}
  \centering
  \includegraphics[width=1\linewidth]{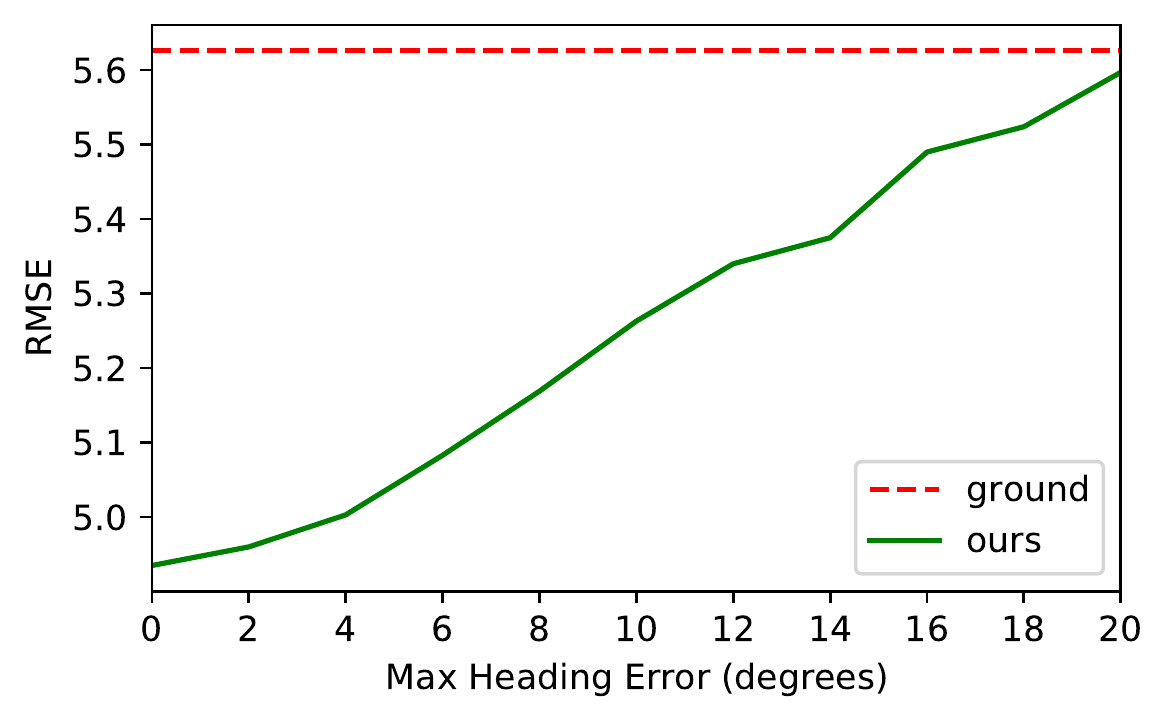}
  \caption{Evaluating performance with increasing orientation error.
  Even with significant noise, our approach outperforms the
  ground-only baseline.}
  \label{fig:noise}
\end{figure}

As our approach relies on geospatial context, we explore the ability
of our method to handle increasing levels of error in geo-orientation.
Note that as the HoliCity~\cite{zhou2020holicity} dataset has non-zero
alignment error, previous results already demonstrate this to a
degree. Since high-end systems can achieve position accuracy on the
order of centimeters~\cite{gps}, we assume accurate geolocation and
focus our attention on orientation. Specifically, we follow the
findings of Kok et al.~\cite{kok2017using} who demonstrate that it is
generally easier to obtain accurate roll and pitch estimates from
inertial sensors than it is to obtain accurate heading (yaw)
estimates. We evaluate our approach by adding increasing levels of
maximum heading error, $\theta$, at inference, by sampling uniformly
on the interval $[-\theta, \theta]$. Note that the average error in
this scenario is approximately $\frac{\theta}{2}$. Intuitively,
performance should decrease as orientation error increases.
\figref{noise} shows the results of this experiment. Our approach
still outperforms the ground-only baseline even with significant added
noise. Additionally, in the supplemental material we demonstrate how
components of our approach can be used to refine geo-orientation.

\subsection{Application: Calibrating Self-Supervised Monocular Approaches}

In this section, we demonstrate the potential of our method to be used
as a tool for calibrating self-supervised depth estimation approaches.
As discussed previously, self-supervised monocular methods can only
estimate depth up to an unknown scale and a scaling factor must be
computed to align predictions. Recent work has highlighted that using
the ground-truth to compute this scaling factor is not a practical
solution~\cite{mccraith2020calibrating}. We begin by investigating the
impact this scaling step has on performance by analyzing a recent
state-of-the-art self-supervised approach,
Monodepth2~\cite{godard2019digging}, using the KITTI depth benchmark.

Though Monodepth2 only predicts depth up to an unknown scale, depth
predictions are constrained to a range of [0, 100] meters (for KITTI)
by passing the final logits through a sigmoid activation and scaling
by a fixed max depth value. To align predictions, median scaling is
used, where the scaling factor for each image is computed from the
ratio of the median predicted values and median ground-truth values
(considering only pixels inside the depth cap).
\tabref{kitti_monodepth} shows results for Monodepth2 with and without
median scaling. For this experiment, the depth cap is set to 80m as is
typical for KITTI. To generate these results, we use a pretrained
model and evaluation scripts made available by the authors. As
observed, median scaling has a drastic impact on performance, with the
average root-mean-square error (meters) increasing by almost a factor
of four when it is disabled.

Next, we evaluate the ability of our approach to be used as a
calibration tool. For this experiment, we use the HoliCity-Overhead
dataset as overhead imagery and height data are not available for
KITTI. Note that retraining Monodepth2 on HoliCity is not possible due
to the lack of image sequences. Using the same process outlined above
and the same pretrained model, we replace the ground-truth depth
values in median scaling with our intermediate representation of
scale. \tabref{holicity_monodepth} shows results for three different
scenarios: with median scaling disabled, median scaling using the
ground truth, and median scaling using the depth from the voxelized
overhead height map as in our approach. As observed, when ground-truth
is not available our approach drastically improves results compared to
no scaling. 

\begin{table}
  \centering
  \caption{Evaluating Monodepth2~\cite{godard2019digging} on KITTI.}
  \begin{tabular}{@{}lcccc@{}}
    \toprule
    & RMSE & RMSE log \\
    \hline
    no scaling & 19.176 & 3.459 \\
    median scaling (ground truth) & 4.863 & 0.193 \\
    \bottomrule
  \end{tabular}
  \label{tbl:kitti_monodepth}
\end{table}

\begin{table}
  \centering
  \caption{Evaluating Monodepth2~\cite{godard2019digging} on HoliCity.}
  \begin{tabular}{@{}lcc@{}}
    \toprule
    & RMSE & RMSE log \\
    \hline
    no scaling                    & 17.555 & 3.054 \\
    median scaling (overhead)     & 15.743 & 1.138 \\
    median scaling (ground truth) & 14.105 & 1.064 \\
    \bottomrule
    \end{tabular}
    \label{tbl:holicity_monodepth}
\end{table}

\section{Conclusion}

We explored a new problem, {\em geo-enabled depth estimation}, in
which the geospatial context of a query image is leveraged to improve
depth estimation. Our key insight was that overhead imagery can serve
as a valuable source of information about the scale of the scene.
Taking advantage of this, we proposed an end-to-end architecture that
integrates geospatial context by first generating an intermediate
representation of the scale of the scene from an estimated (or known)
height map and then fusing it inside of a segmentation architecture
that operates on a ground-level image. An extensive evaluation shows
that our method significantly reduces error compared to baselines,
especially when considering much greater distances than existing
evaluation benchmarks. Ultimately our hope is that this work
demonstrates that existing depth estimation techniques can benefit
when geospatial context is available.

{\small
\bibliographystyle{ieee_fullname}
\bibliography{biblio}
}

\newpage
\null
\vskip .375in
\twocolumn[{%
  \begin{center}
    \textbf{\Large Supplemental Material : \\ Augmenting Depth Estimation with Geospatial Context}
  \end{center}
  \vspace*{24pt}
}]
\setcounter{section}{0}
\setcounter{equation}{0}
\setcounter{figure}{0}
\setcounter{table}{0}
\makeatletter
\renewcommand{\theequation}{S\arabic{equation}}
\renewcommand{\thefigure}{S\arabic{figure}}
\renewcommand{\thetable}{S\arabic{table}}

This document contains additional details and experiments related to
our methods.

\section{HoliCity-Overhead Dataset}

\begin{figure}

  \centering

  \fbox{\includegraphics[width=1\linewidth]{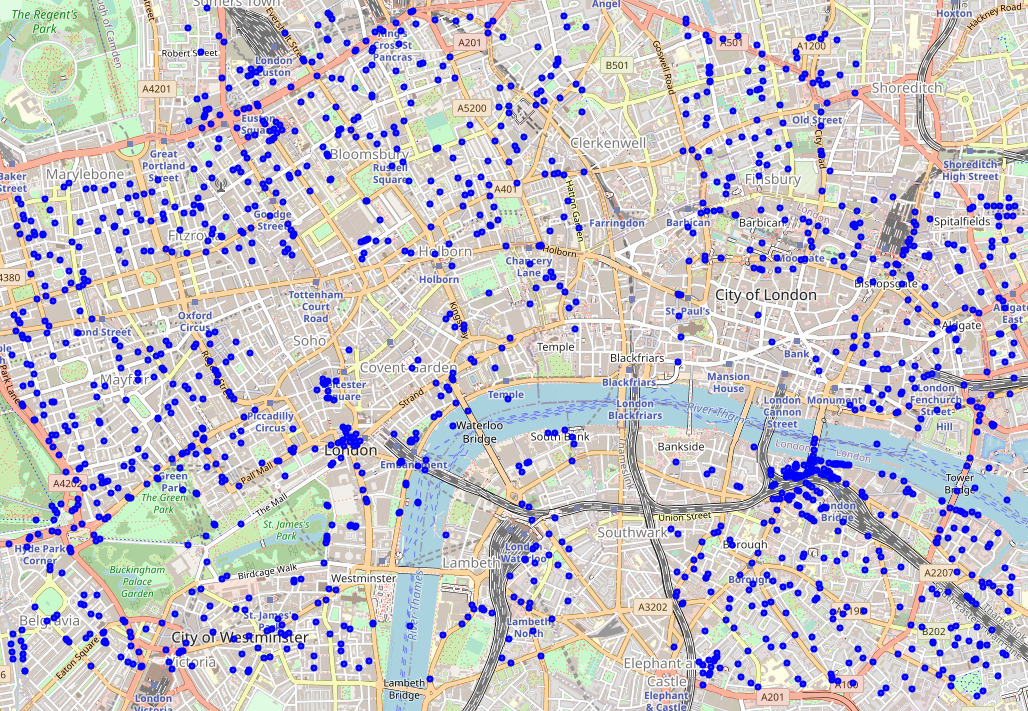}}

  \caption{Visualizing the coverage of the
  HoliCity~\cite{zhou2020holicity} dataset. The locations of the
  ground-level panoramas are shown as blue dots (subsampled).}

  \label{fig:coverage}
\end{figure}

\begin{figure}

  \centering

  \setlength\tabcolsep{1pt}

  \begin{tabular}{ccc}

    zoom 18 & zoom 17 & zoom 16 \\

    \includegraphics[width=.32\linewidth]{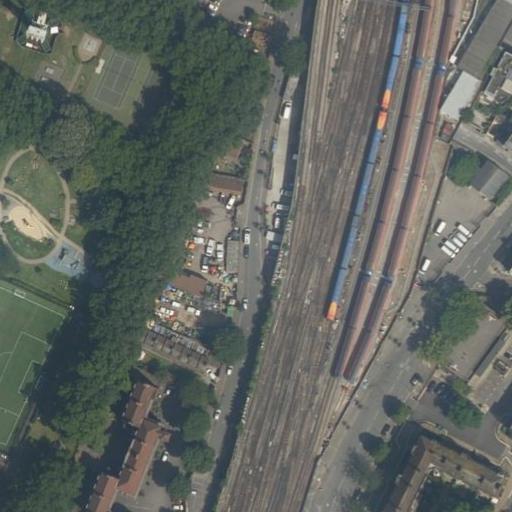} &
    \includegraphics[width=.32\linewidth]{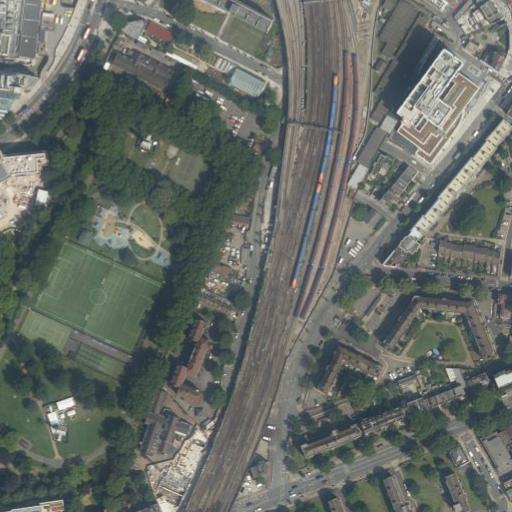} &
    \includegraphics[width=.32\linewidth]{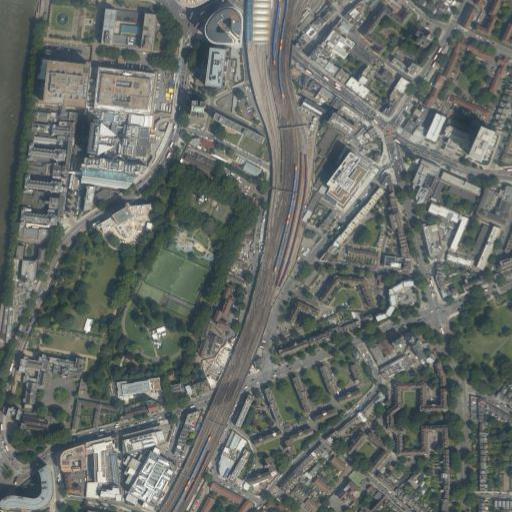} \\    
        
    \includegraphics[width=.32\linewidth]{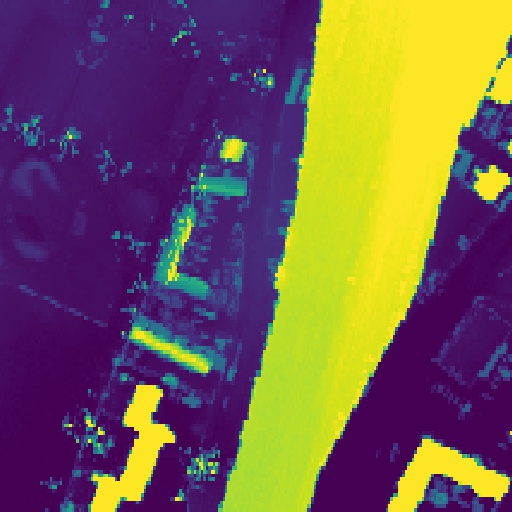} &
    \includegraphics[width=.32\linewidth]{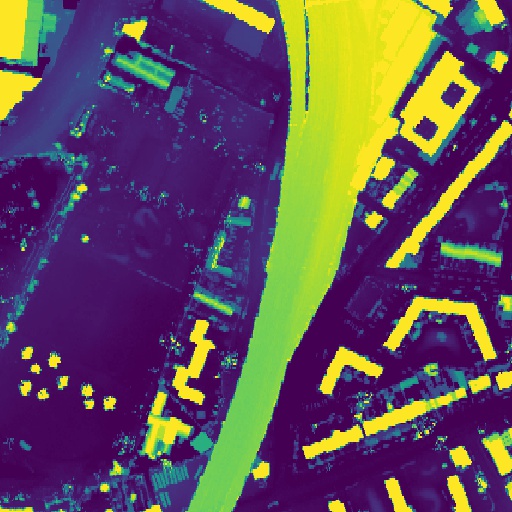} &
    \includegraphics[width=.32\linewidth]{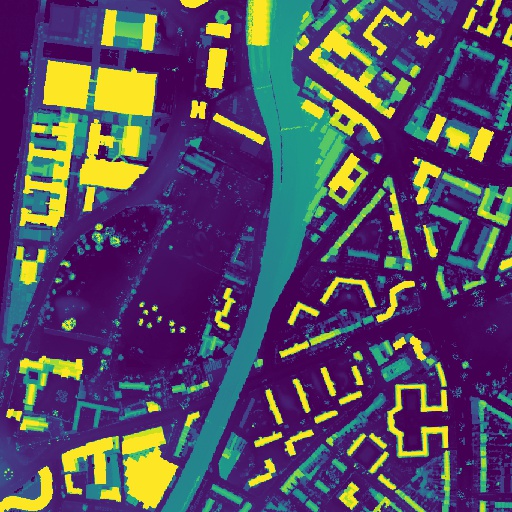} \\
    
  \end{tabular}

  \caption{Example data from the HoliCity-Overhead dataset at
  different zoom levels (i.e., varying ground sample distance). (top)
  Overhead image and (bottom) height map from composite digital
  surface model.}

  \label{fig:zoom}
\end{figure}

We introduced the HoliCity-Overhead dataset, an extension of the the
HoliCity~\cite{zhou2020holicity} dataset that includes overhead
imagery and height data. \figref{coverage} visualizes the coverage of
the original dataset in downtown London, UK. \figref{zoom} shows
example overhead images and aligned height map pairs (each image of
size $512 \times 512$) at the different zoom levels we collected
(i.e., varying ground sample distance or ground resolution). 

\section{Extended Results}

We present additional evaluation of our methods using the
HoliCity-Overhead dataset.

\subsection{Height Estimation}

\begin{figure*}[!ht]

  \centering

  \setlength\tabcolsep{1pt}

  \begin{tabular}{ccccc}

    \raisebox{1.2\height}{\rotatebox{90}{image}}
    \includegraphics[width=.18\linewidth]{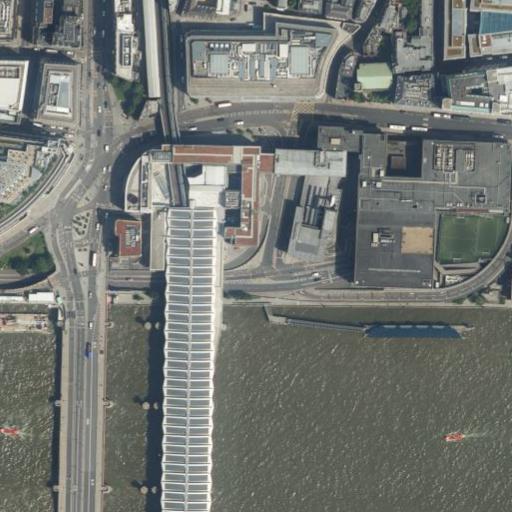} &
    \includegraphics[width=.18\linewidth]{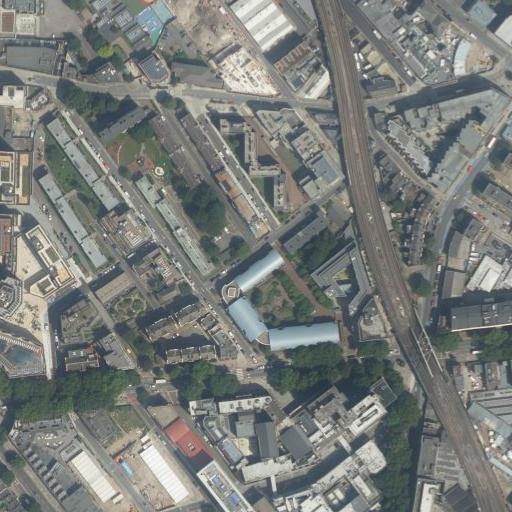} &
    \includegraphics[width=.18\linewidth]{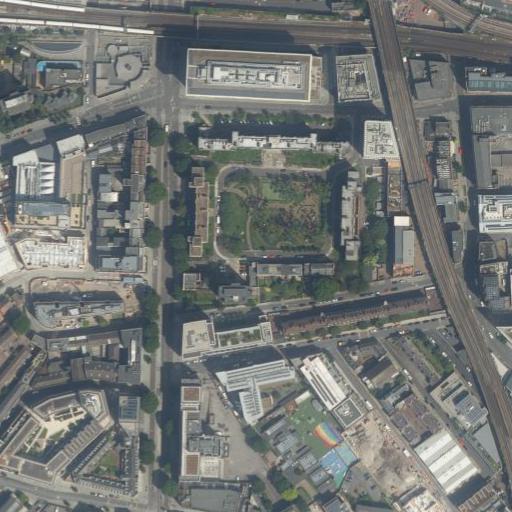} &
    \includegraphics[width=.18\linewidth]{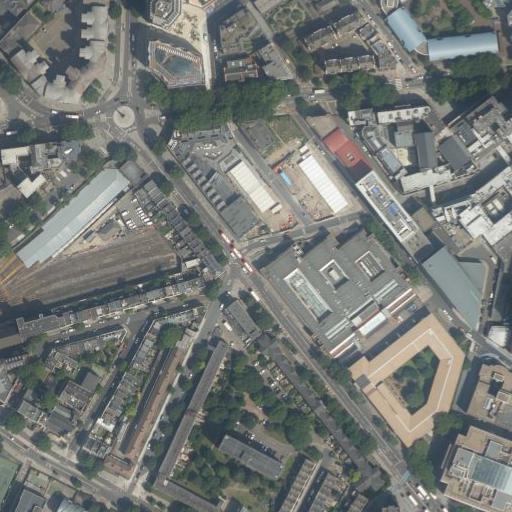} &
    \includegraphics[width=.18\linewidth]{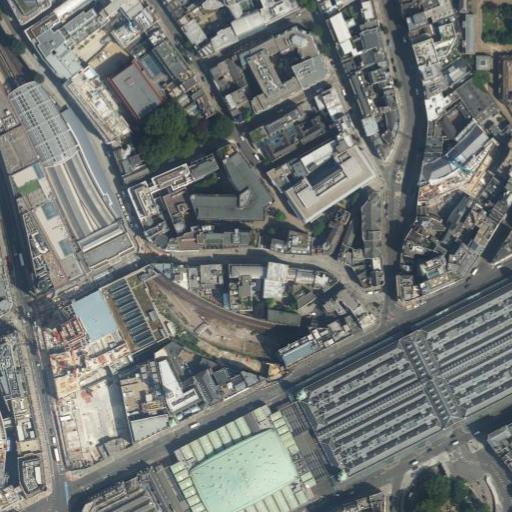} \\
    
    \raisebox{1.55\height}{\rotatebox{90}{label}}
    \includegraphics[width=.18\linewidth]{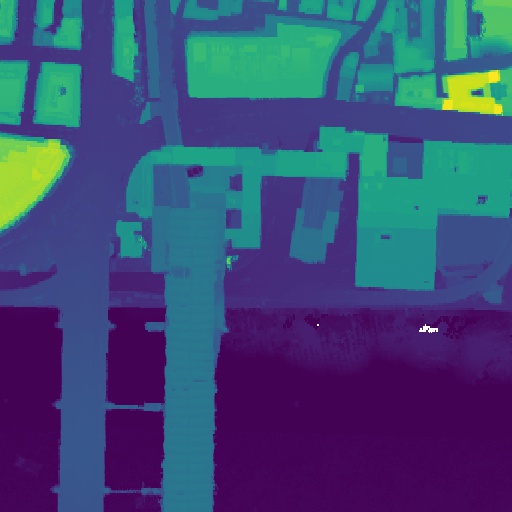} &
    \includegraphics[width=.18\linewidth]{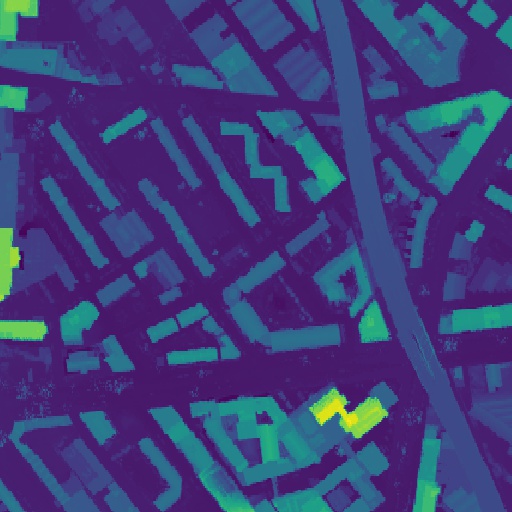} &
    \includegraphics[width=.18\linewidth]{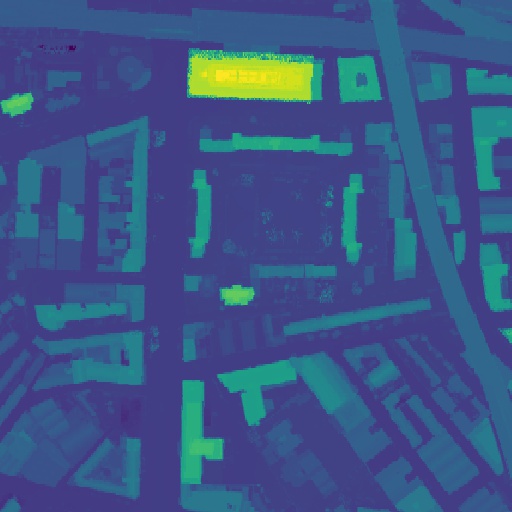} &
    \includegraphics[width=.18\linewidth]{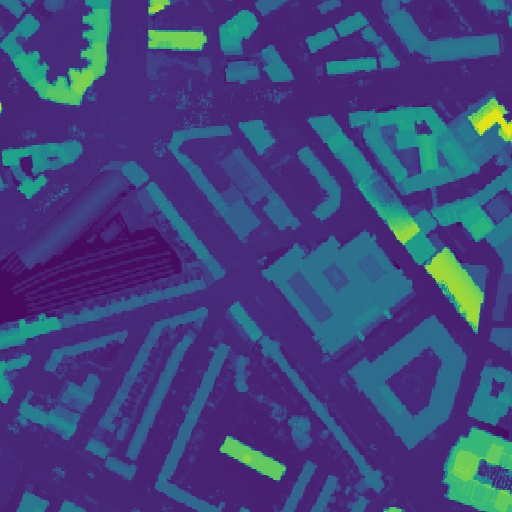} &
    \includegraphics[width=.18\linewidth]{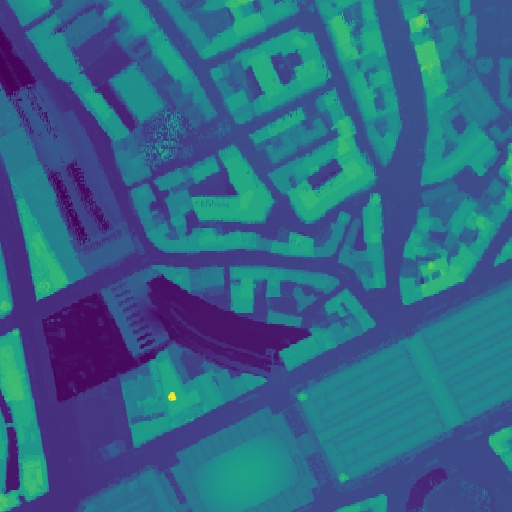} \\
    
    \raisebox{.55\height}{\rotatebox{90}{prediction}}
    \includegraphics[width=.18\linewidth]{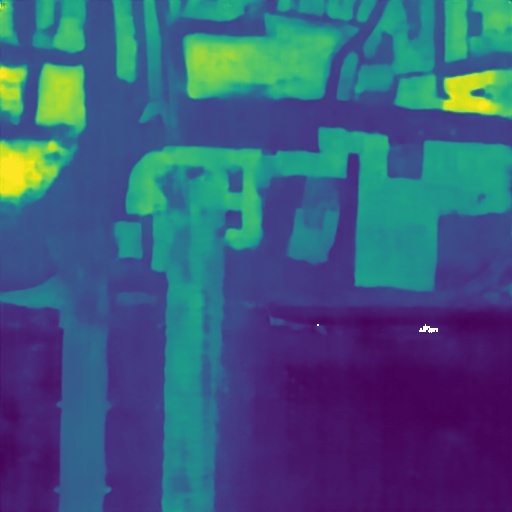} &   
    \includegraphics[width=.18\linewidth]{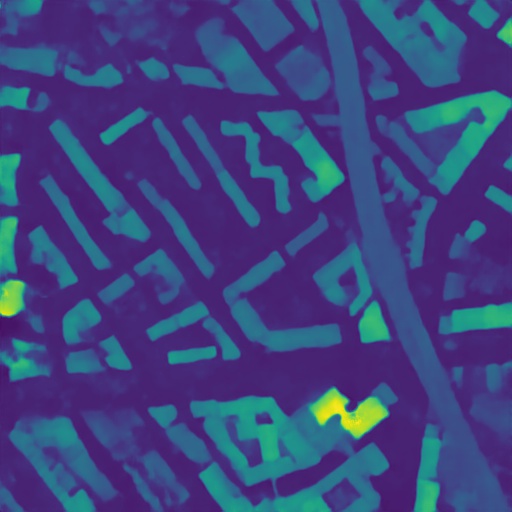} &  
    \includegraphics[width=.18\linewidth]{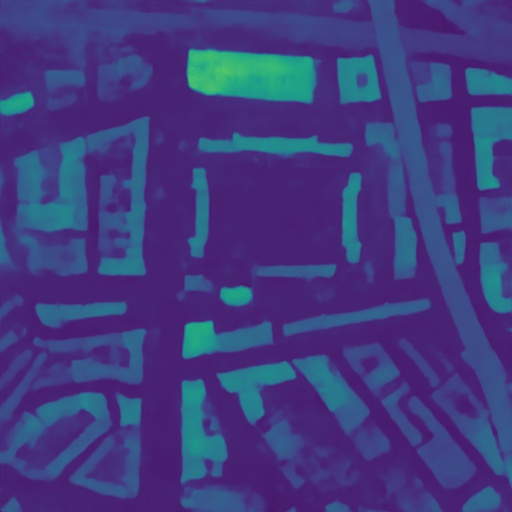} &  
    \includegraphics[width=.18\linewidth]{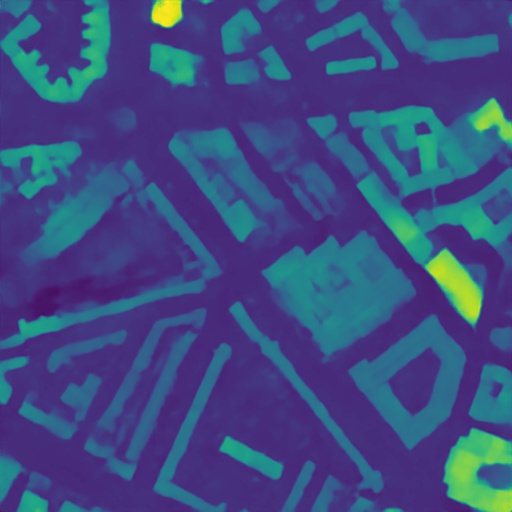} &  
    \includegraphics[width=.18\linewidth]{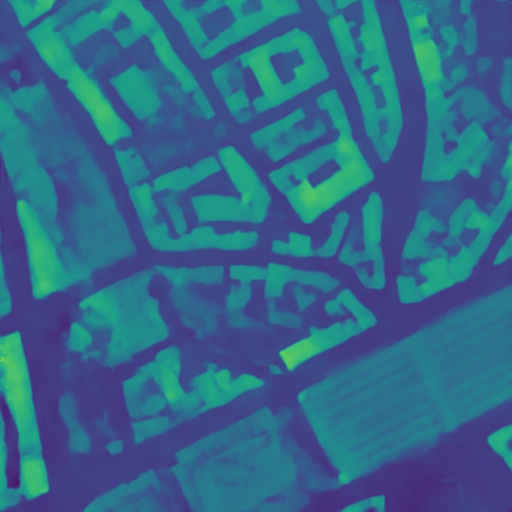} \\
    
  \end{tabular}

  \caption{Example height maps generated by our approach where yellow
  represents larger values.}

  \label{fig:height}
\end{figure*}

\tabref{results_height} shows results for height estimation, generated
using our method (Section 4 of the main paper). For this experiment we
used the zoom level 17 imagery, which equates to a ground sample
distance of approximately .74 meters per pixel over London. Though
height estimation is only an intermediate task of our network and not
the primary objective, our approach achieves good performance in
several metrics. \figref{height} shows example height maps generated
using our approach alongside the ground truth, with yellow indicating
larger values.

\setlength{\fboxsep}{0pt}
\setlength{\fboxrule}{1px}

\begin{table}[!h]
  \centering
  \caption{Height estimation results on HoliCity-Overhead.}
  \begin{tabular}{@{}lccc@{}}
    \toprule
    & MAE & RMSE & RMSE log \\
    \hline
    ours (height est.)               & 3.333 & 5.225  & 0.470 \\
    \bottomrule
    \end{tabular}
    \label{tbl:results_height}
\end{table}

\subsection{Scale Factor Analysis}

\begin{figure}

  \centering

  \includegraphics[width=1\linewidth]{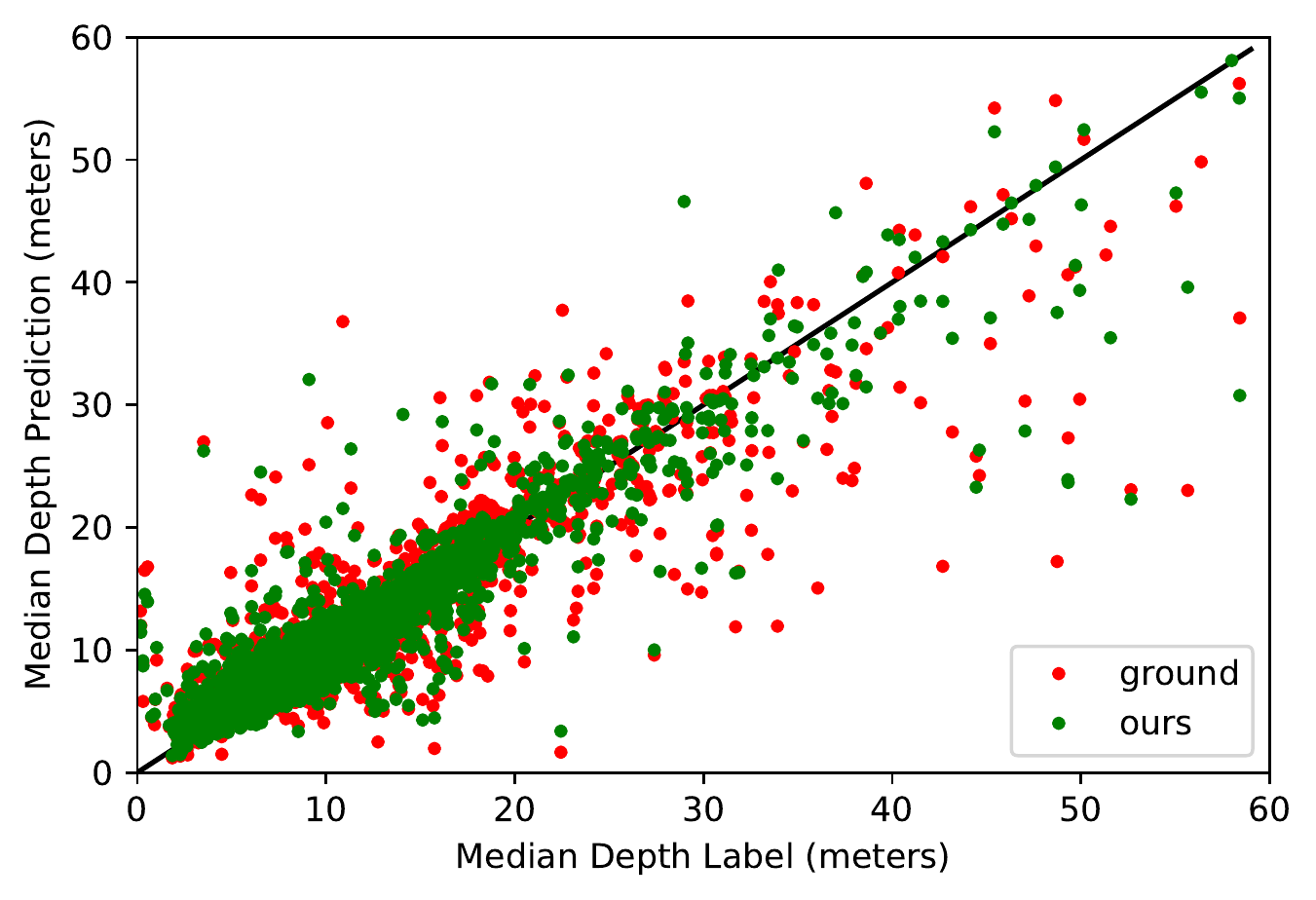}

  \caption{Visualizing median depth of ground-truth versus median
  depth of prediction (the ratio of which is the scale factor for
  median scaling) for our method versus a baseline. Each dot
  corresponds to an image in the HoliCity-Overhead evaluation set. Our
  method is generally closer to the diagonal, indicating better
  performance.}

  \label{fig:median}
\end{figure}

Though our method requires no computation of a scaling factor at
inference, we can analyze the performance of our method in terms of
scale. \figref{median} shows a scatter plot (per image) where the
x-axis is median depth from ground-truth and the y-axis is median
depth from prediction. Note that in median scaling, the scaling factor
is estimated as the ratio of these two quantities. For this
visualization, points closer to the diagonal indicate better alignment
in scale. As observed, our approach is generally closer to the
diagonal when compared against the ground-only baseline.

\begin{figure*}
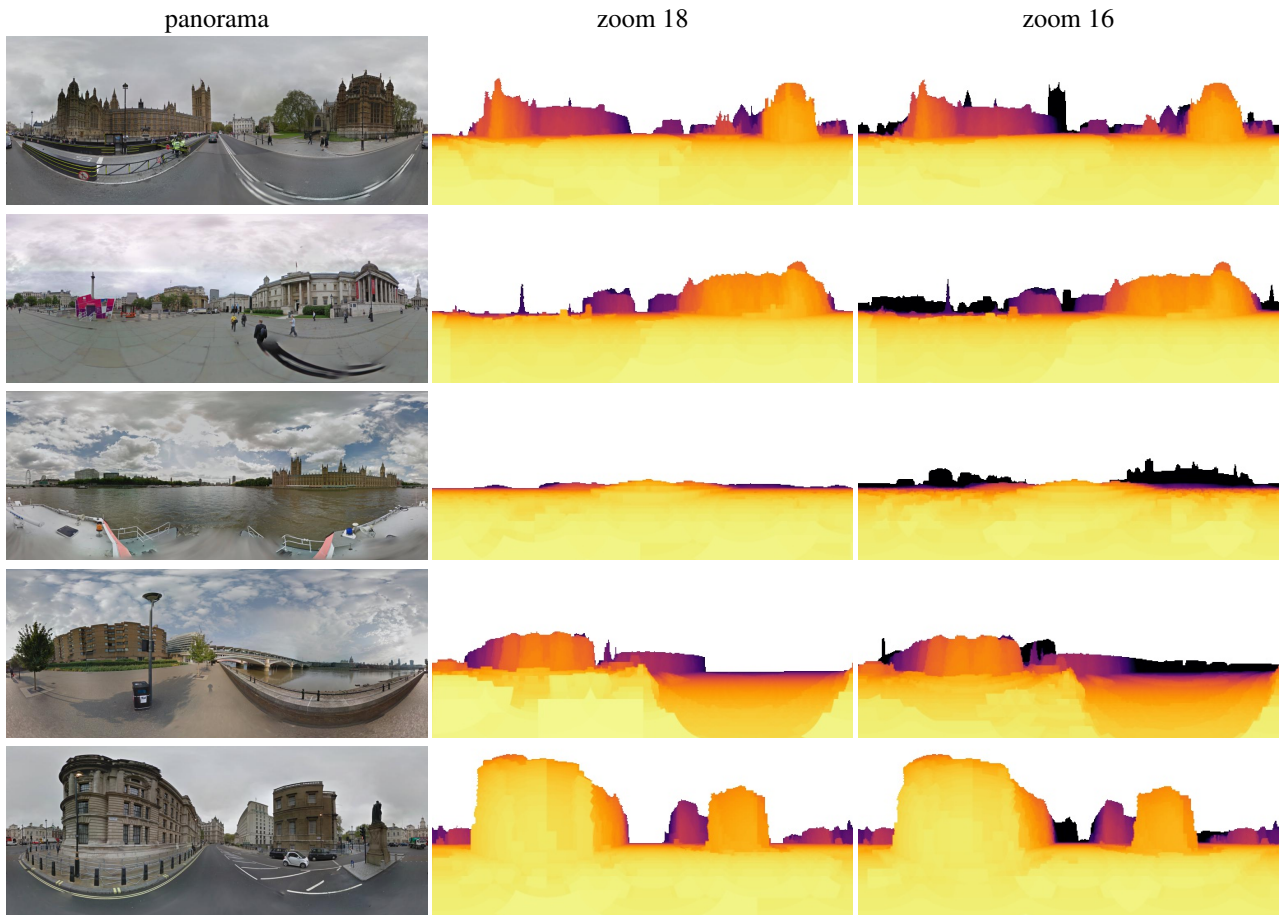


  \centering

  \setlength\tabcolsep{1pt}

  \begin{tabular}{cccc}

    panorama & zoom 18 & zoom 16 \\
    
    \adjincludegraphics[trim=0 {.15\height} 0 {.05\height},clip,width=.32\linewidth]{supplemental/zoom/1152_pano} &
    \adjincludegraphics[trim=0 {.15\height} 0 {.05\height},clip,width=.32\linewidth]{supplemental/zoom/1152_depth_pano_17} &
    \adjincludegraphics[trim=0 {.15\height} 0 {.05\height},clip,width=.32\linewidth]{supplemental/zoom/1152_depth_pano_15} \\
    
    \adjincludegraphics[trim=0 {.15\height} 0 {.05\height},clip,width=.32\linewidth]{supplemental/zoom/2952_pano} &
    \adjincludegraphics[trim=0 {.15\height} 0 {.05\height},clip,width=.32\linewidth]{supplemental/zoom/2952_depth_pano_17} &
    \adjincludegraphics[trim=0 {.15\height} 0 {.05\height},clip,width=.32\linewidth]{supplemental/zoom/2952_depth_pano_15} \\
    
    \adjincludegraphics[trim=0 {.15\height} 0 {.05\height},clip,width=.32\linewidth]{supplemental/zoom/3416_pano} &
    \adjincludegraphics[trim=0 {.15\height} 0 {.05\height},clip,width=.32\linewidth]{supplemental/zoom/3416_depth_pano_17} &
    \adjincludegraphics[trim=0 {.15\height} 0 {.05\height},clip,width=.32\linewidth]{supplemental/zoom/3416_depth_pano_15} \\
    
    \adjincludegraphics[trim=0 {.15\height} 0 {.05\height},clip,width=.32\linewidth]{supplemental/zoom/5445_pano} &
    \adjincludegraphics[trim=0 {.15\height} 0 {.05\height},clip,width=.32\linewidth]{supplemental/zoom/5445_depth_pano_17} &
    \adjincludegraphics[trim=0 {.15\height} 0 {.05\height},clip,width=.32\linewidth]{supplemental/zoom/5445_depth_pano_15} \\
    
    \adjincludegraphics[trim=0 {.15\height} 0 {.05\height},clip,width=.32\linewidth]{supplemental/zoom/2008_pano} &
    \adjincludegraphics[trim=0 {.15\height} 0 {.05\height},clip,width=.32\linewidth]{supplemental/zoom/2008_depth_pano_17} &
    \adjincludegraphics[trim=0 {.15\height} 0 {.05\height},clip,width=.32\linewidth]{supplemental/zoom/2008_depth_pano_15} \\

  \end{tabular}

  \caption{Visualizing the impact of ground sample distance. Starting
  from lower zoom levels (higher ground sample distance) increases
  spatial coverage, enabling capture of objects farther away in the
  synthetic depth panoramas. Yellow (black) values indicate smaller
  (larger) depths.}

  \label{fig:zoom_out}
\end{figure*}

\begin{figure*}

  \centering

  \setlength\tabcolsep{1pt}

  \begin{tabular}{ccc|c}

    image & ground & ours & label \\
    
    \includegraphics[width=.24\linewidth]{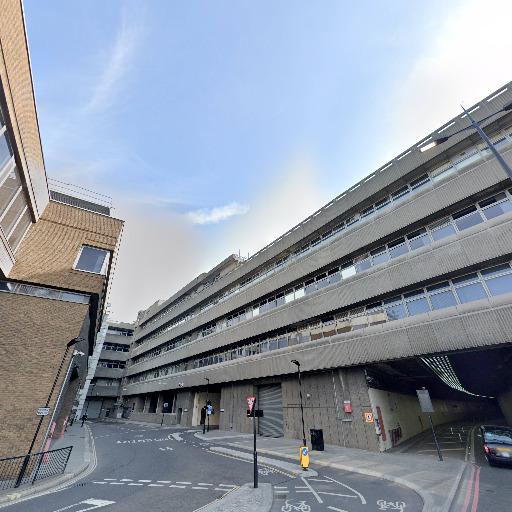} &
    \includegraphics[width=.24\linewidth]{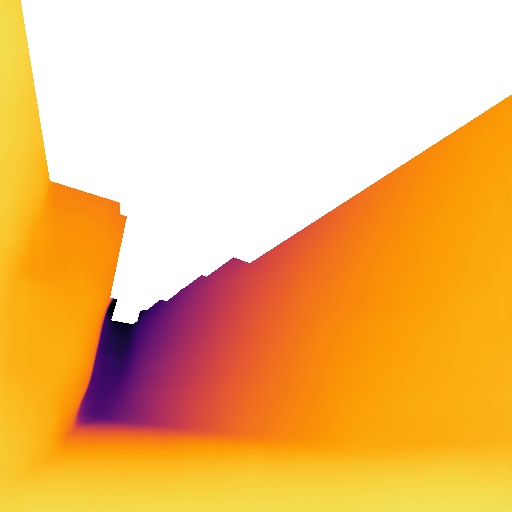} &
    \includegraphics[width=.24\linewidth]{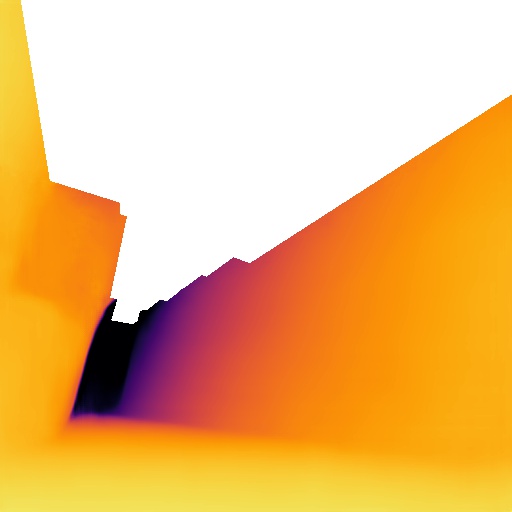} &
    \includegraphics[width=.24\linewidth]{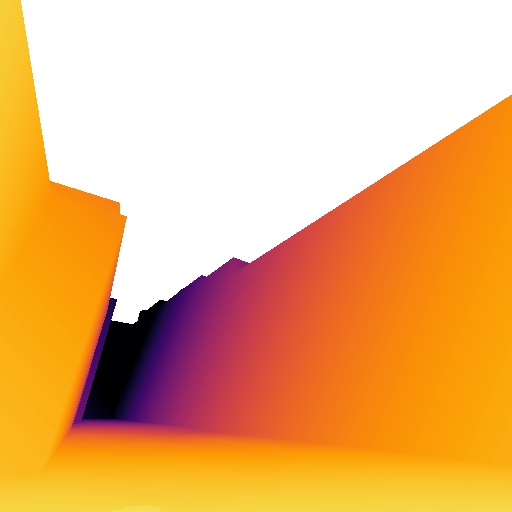} \\
    
    \includegraphics[width=.24\linewidth]{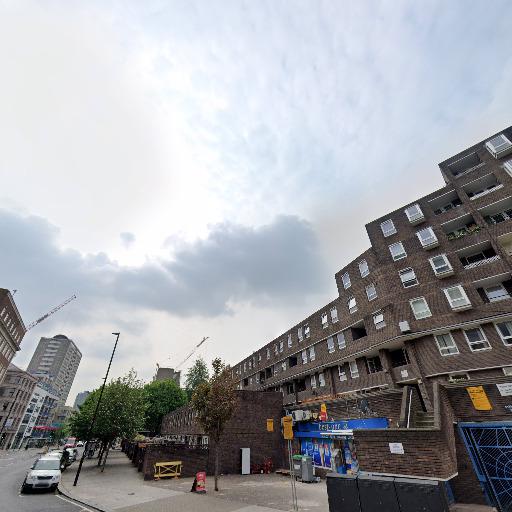} &
    \includegraphics[width=.24\linewidth]{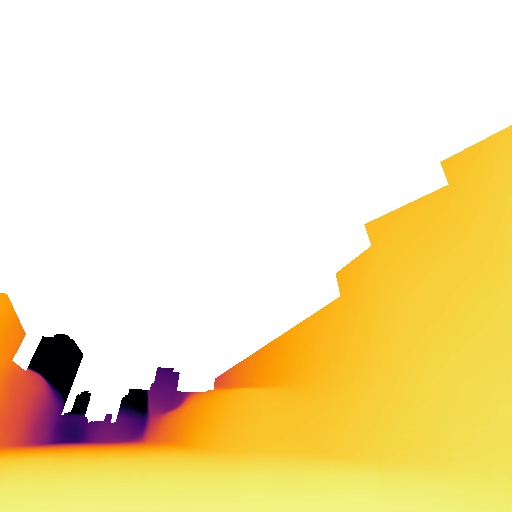} &
    \includegraphics[width=.24\linewidth]{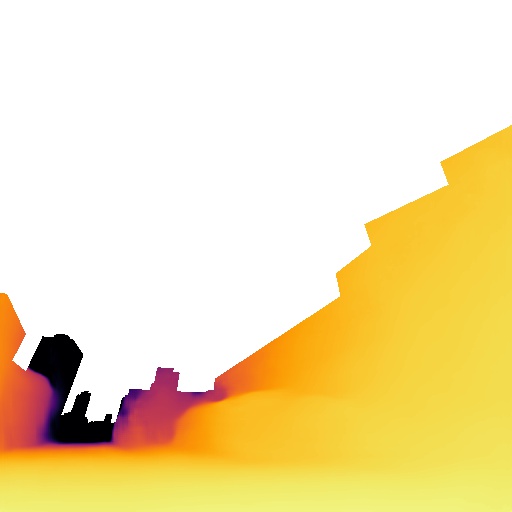} &
    \includegraphics[width=.24\linewidth]{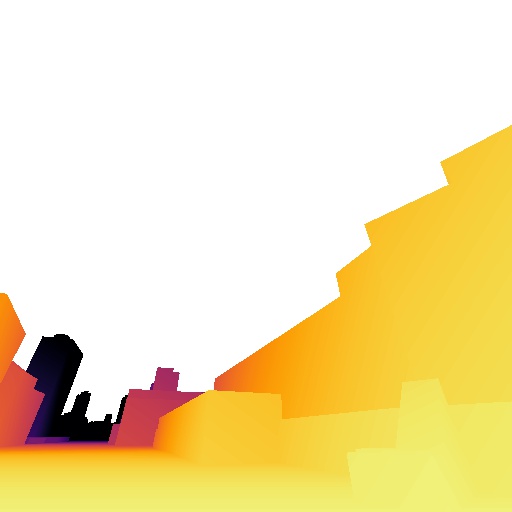} \\
    
    \includegraphics[width=.24\linewidth]{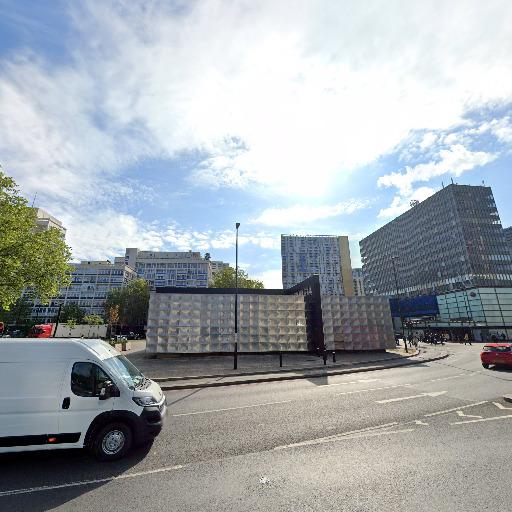} &
    \includegraphics[width=.24\linewidth]{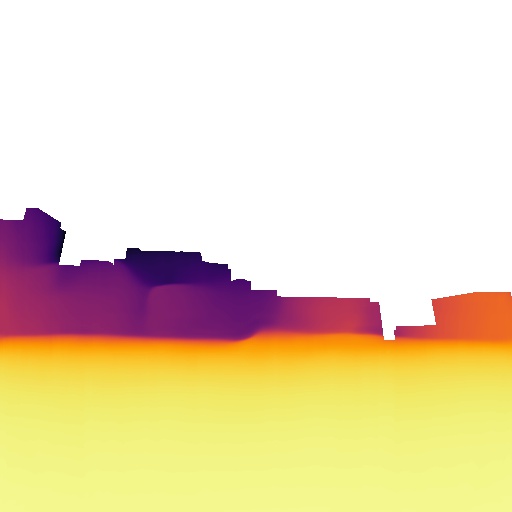} &
    \includegraphics[width=.24\linewidth]{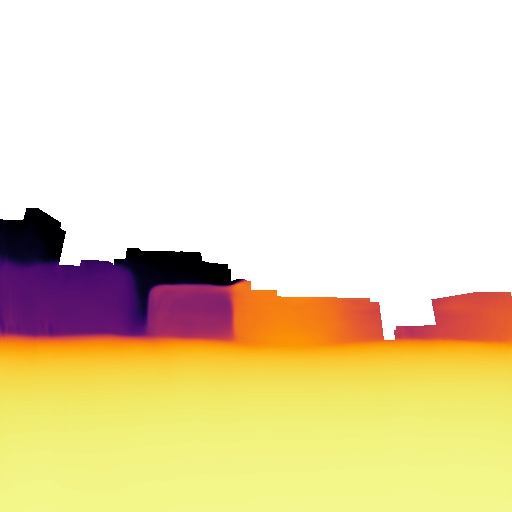} &
    \includegraphics[width=.24\linewidth]{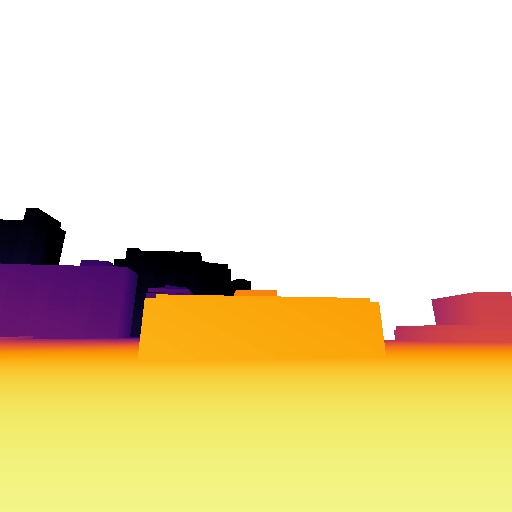} \\
    
    \includegraphics[width=.24\linewidth]{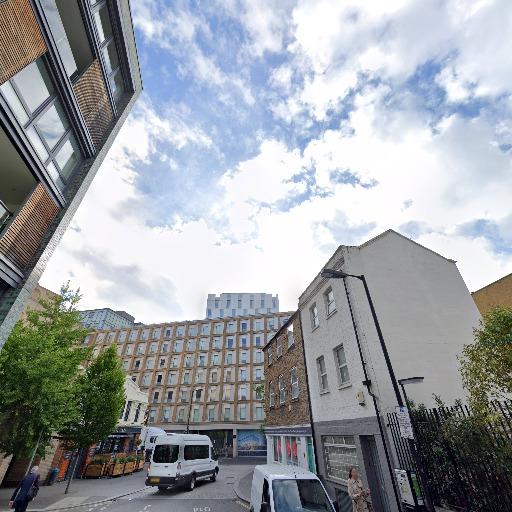} &
    \includegraphics[width=.24\linewidth]{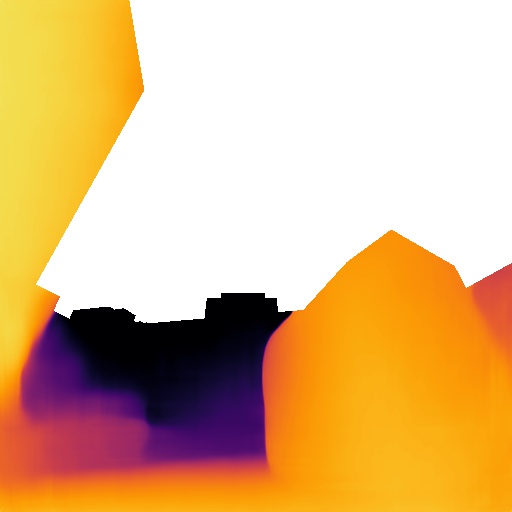} &
    \includegraphics[width=.24\linewidth]{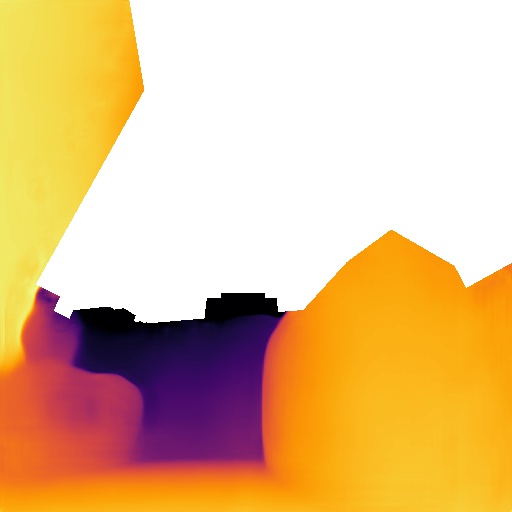} &
    \includegraphics[width=.24\linewidth]{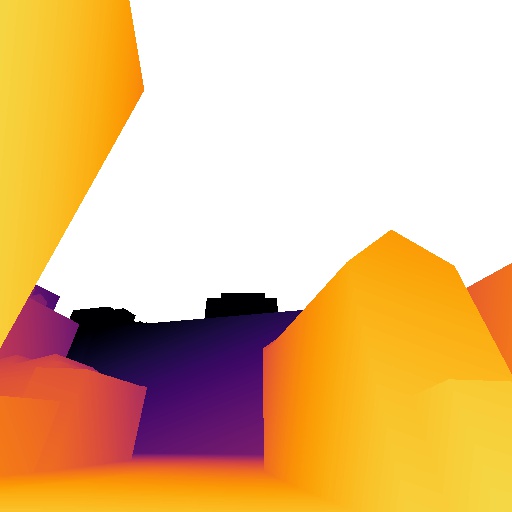} \\
    
  \end{tabular}

  \caption{Qualitative results of our method compared to the
  ground-only baseline. Our approach, which integrates geospatial
  context, better captures the scale of the scene. Yellow (black)
  values indicate smaller (larger) depths.}

  \label{fig:results}
\end{figure*}

\subsection{Impact of Ground Sample Distance}

As shown in Section 5.2 of the main paper, starting from an overhead
height map with larger ground sample distance leads to better
performance when integrating geospatial context. This makes sense
because a larger ground sample distance, but same size image, has
greater spatial coverage. Here we visualize the impact of this on the
generated synthetic depth panoramas. For this experiment, we use the
ground-truth height map data contained in the HoliCity-Overhead
dataset. \figref{zoom_out} visualizes the results. As observed,
synthetic depth panoramas generated from zoom level 16 data (approx.
1.5 meters per pixel) contain objects, such as buildings, that are
farther away than in zoom level 18 (approx. .4 meters per pixel).

\subsection{Qualitative Results}

Finally, in \figref{results} we show results generated by our method
alongside results from the ground-only baseline. As observed, our
approach that integrates geospatial context is better able to capture
the scale of the scene. For example, in the first row, the ground-only
baseline significantly underestimates the maximum depth along the road
compared to our approach.

\section{Application: Estimating Geo-Orientation}

\begin{figure*}

  \centering
  \begin{tabular}{cc}
    \includegraphics[trim=0 100 0 50,clip,width=0.5\linewidth]{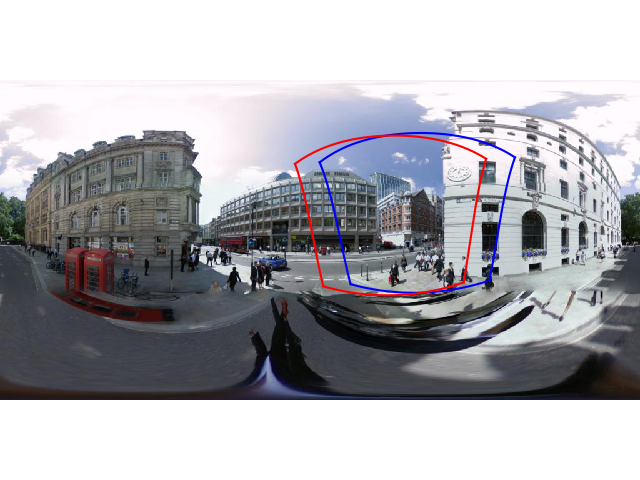} & 
    \includegraphics[trim=0 100 0 50,clip,width=0.5\linewidth]{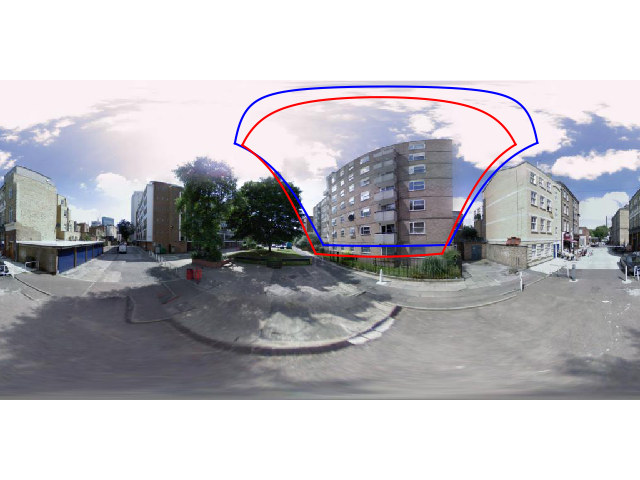} \\
    
    \includegraphics[trim=0 100 0 50,clip,width=0.5\linewidth]{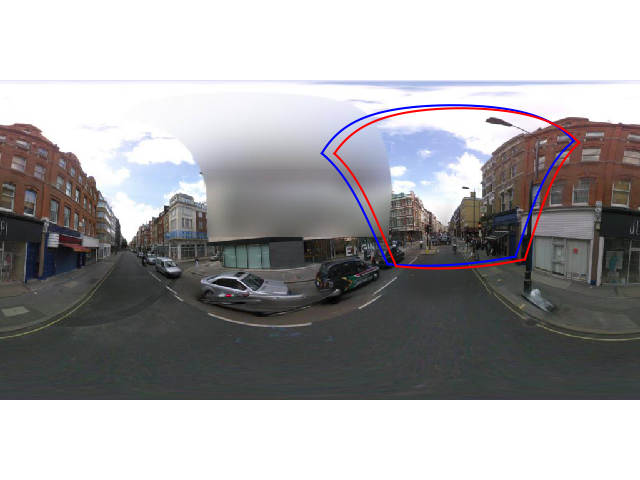} & 
    \includegraphics[trim=0 100 0 50,clip,width=0.5\linewidth]{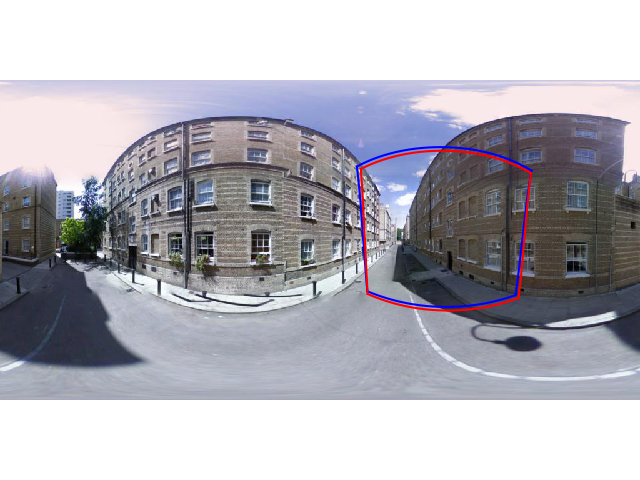} \\
    
    \includegraphics[trim=0 100 0 50,clip,width=0.5\linewidth]{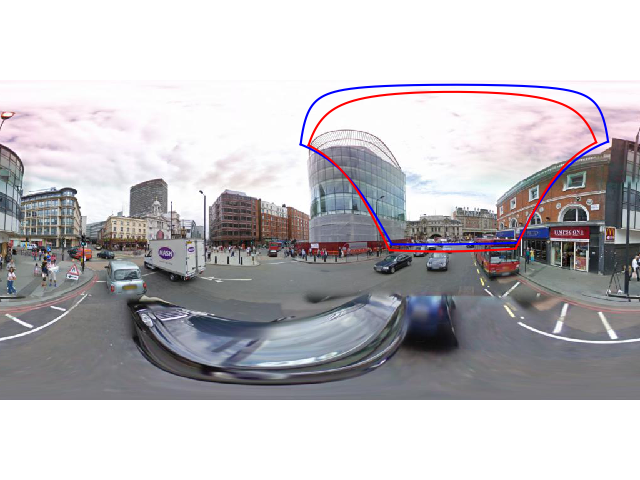} & 
    \includegraphics[trim=0 100 0 50,clip,width=0.5\linewidth]{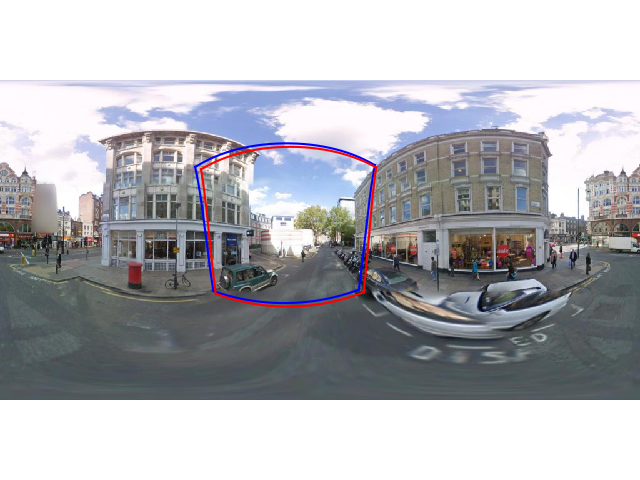} \\
    
    \includegraphics[trim=0 100 0 50,clip,width=0.5\linewidth]{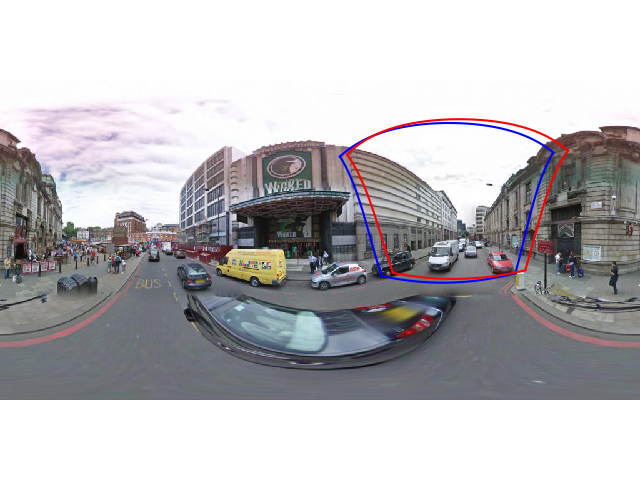} & 
    \includegraphics[trim=0 100 0 50,clip,width=0.5\linewidth]{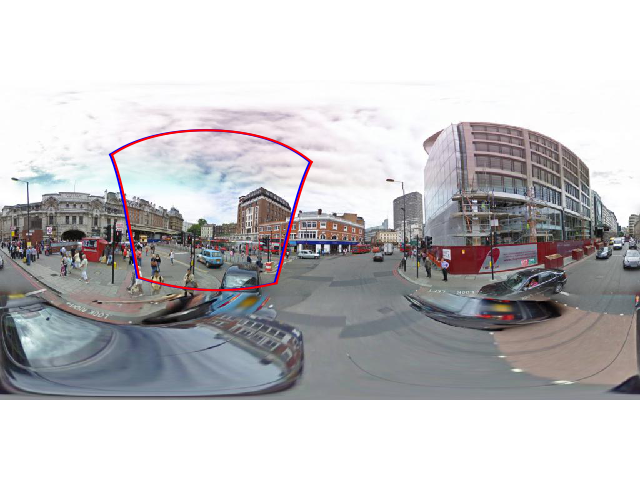} \\
  \end{tabular}
  
  \caption{Example results from orientation estimation. The
  perspective image boundary corresponding to the true orientation is
  shown in blue and our registration result is shown in red.}

  \label{fig:registration}
\end{figure*}

We show that our intermediate representation of scale (in the form of
a synthetic depth panorama) can be used for a variety of applications,
including orientation estimation. For this experiment, we use the
overhead height map at zoom level 16. Given a query ground-level depth
map with known geolocation but unknown orientation, we first generate
the synthetic depth panorama from a co-located height map using our
approach. We then perform a grid search over yaw/pitch parameters at
one degree intervals, extracting the corresponding perspective depth
cutout, and comparing to the query depth image. Each orientation is
assigned a score using mean absolute error between the two depth maps,
selecting the lowest error configuration as our prediction. Example
registration results are shown in \figref{registration}. The
ground-truth perspective cutout boundary is shown in blue and the
result of our registration technique is shown in red. Despite this
simple approach, the estimated orientations are quite accurate.

\end{document}